\PassOptionsToPackage{table}{xcolor}
\documentclass[10pt]{article} %
\usepackage[preprint]{tmlr}

\usepackage{amsmath,amsfonts,bm}

\def\Secref#1{Section~\ref{#1}}

\def\eqref#1{equation~\ref{#1}}

\def\1{\bm{1}}

\def\vi{{\bm{i}}}

\def\vz{{\bm{z}}}

\def\mI{{\bm{I}}}

\DeclareMathAlphabet{\mathsfit}{\encodingdefault}{\sfdefault}{m}{sl}
\SetMathAlphabet{\mathsfit}{bold}{\encodingdefault}{\sfdefault}{bx}{n}

\def\gN{{\mathcal{N}}}

\def\gT{{\mathcal{T}}}
\def\gU{{\mathcal{U}}}

\usepackage{url}
\usepackage{colors}
\usepackage[pagebackref=true,
            breaklinks=true,
            colorlinks,
            linkcolor=mycitecolor,
            urlcolor=mycitecolor,
            anchorcolor=mycitecolor,
            citecolor=mycitecolor,
            bookmarks=false]{hyperref}
\usepackage{graphicx}
\usepackage[caption=true,font=footnotesize, justification= centering]{subfig}
\usepackage{amsmath}
\usepackage{amssymb}
\usepackage{multirow}
\usepackage{makecell}
\usepackage{bbm}
\usepackage{enumitem}
\usepackage{mathtools}
\usepackage{amsthm}

\usepackage{wrapfig}
\usepackage[ruled]{algorithm2e}
\SetKwComment{tcp}{$\#$ }{}

\SetCommentSty{mycommfont}

\usepackage{enumitem}
\usepackage{tikz}
\usepackage{ctable}
\usepackage[capitalize,noabbrev]{cleveref}
\def\ie{\emph{i.e.}}

\usepackage{eqparbox}

\definecolor{bg_blue}{HTML}{e6efff}

\title{SASSL: Enhancing Self-Supervised Learning via\\ Neural Style Transfer}

\author{\name Renan A. Rojas-Gomez$^{*}$ \email renanar2@illinois.edu \\
      \addr University of Illinois at Urbana-Champaign
      \AND
      \name Karan Singhal\email karansinghal@google.com \\
      \addr Google Research
      \AND
      \name Ali Etemad \email ali.etemad@queensu.ca\\
      \addr Queen’s University, Canada
      \AND
      \name Alex Bijamov \email abijamov@google.com\\
      \addr Google DeepMind
      \AND
      \name Warren R. Morningstar \email wmorning@google.com\\
      \addr Google DeepMind
      \AND
      \name Philip Andrew Mansfield \email memes@google.com\\
      \addr Google DeepMind\\\\\\
    \normalsize{$^*$Work done during an internship at Google.}
      }

\def\month{11}  %
\def\year{2024} %
\def\openreview{\url{https://openreview.net/forum?id=NxhXtkPYsk}} %

\begin{document}
\maketitle
\begin{abstract}
Existing data augmentation in self-supervised learning, while diverse, fails to preserve the inherent structure of natural images. This results in distorted augmented samples with compromised semantic information, ultimately impacting downstream performance. To overcome this limitation, we propose \emph{SASSL: Style Augmentations for Self Supervised Learning}, a novel data augmentation technique based on Neural Style Transfer. SASSL decouples semantic and stylistic attributes in images and applies transformations exclusively to their style while preserving content, generating diverse samples that better retain semantic information. SASSL boosts top-1 image classification accuracy on ImageNet by up to 2 percentage points compared to established self-supervised methods like MoCo, SimCLR, and BYOL, while achieving superior transfer learning performance across various datasets. Because SASSL can be performed asynchronously as part of the data augmentation pipeline, these performance impacts can be obtained with no change in pretraining throughput.
\end{abstract}

\section{Introduction}
\label{sec:intro}
Data labelling is a challenging and expensive process, which often serves as a barrier to build machine learning models to solve real-world problems. Self-supervised learning (SSL) is an emerging machine learning paradigm that helps to alleviate the challenges of data labelling, by using large corpora of unlabeled data to pretrain models to learn robust and general representations. These representations can be efficiently transferred to downstream tasks, resulting in performant models which can be constructed without access to large pools of labeled data. SSL methods have shown promising results in recent years, matching and in some cases exceeding the performance of bespoke supervised models with small amounts of labelled data.

Given the lack of labels, SSL relies on pretext tasks, \ie, predefined tasks where pseudo-labels can be generated. These include contrastive learning \citep{chen_2020_simple,he_2020_momentum}, clustering \citep{caron_2021_emerging, caron_2020_unsupervised, assran_2022_masked}, and generative modeling \citep{he_2022_masked,devlin_2018_bert}. Many pretext tasks involve training the model to distinguish between different views of the same input and inputs corresponding to different samples. For these tasks, the way input data is augmented is crucial to learn useful invariances and extract robust representations \citep{chen_2020_simple}. While state-of-the-art augmentations incorporate a wide range of color, spectral and spatial transformations, they often disregard the natural structure of an image. As a result, SSL pretraining methods may generate augmented samples with degraded semantic information, and may be less able to capture diverse visual attributes.

To tackle this challenge, we propose \emph{Style Augmentations for Self Supervised Learning (SASSL)}, a novel SSL data augmentation technique based on Neural Style Transfer to generate semantically consistent augmented samples. In contrast to augmentation techniques operating on specific formats (e.g. pixel or spectral domain), SASSL disentangles an image into perceptual (\emph{style}) and semantic (\emph{content}) representations that are learned from data. Applying transformations only to the style of an image while preserving its content, we can generate images with diverse appearance that retain the original semantic properties.

\textbf{Our contributions:}
\begin{itemize}[leftmargin=0.45cm,noitemsep,topsep=0pt]
    \item We propose SASSL, a novel data augmentation technique based on Style Transfer that naturally preserves semantic properties while diversifying style (\cref{sec:proposed}).
    \item We empirically show improved downstream performance on ImageNet \cite{deng_2009_imagenet} by incorporating SASSL in methods such as MoCo, BYOL and SimCLR without hyperparameter tuning (Sections \ref{sec:downstream_task_performance}, \ref{sec:additional_downstream_performance}).
    \item We show SASSL learns stronger representations by measuring their transfer learning capabilities on various datasets. Our method boosts linear probing performance by up to $10\%$ and fine-tuning by up to $6\%$ on out-of-distribution tasks (\cref{sec:transfer_learning_performance}).
\end{itemize}

\begin{figure*}[t]
    \vspace{-0.25 cm}
    \centering
    \subfloat[\textbf{Data augmentation pipeline.}]{\makebox[0.29\textwidth]{\includegraphics[height=0.24\linewidth]{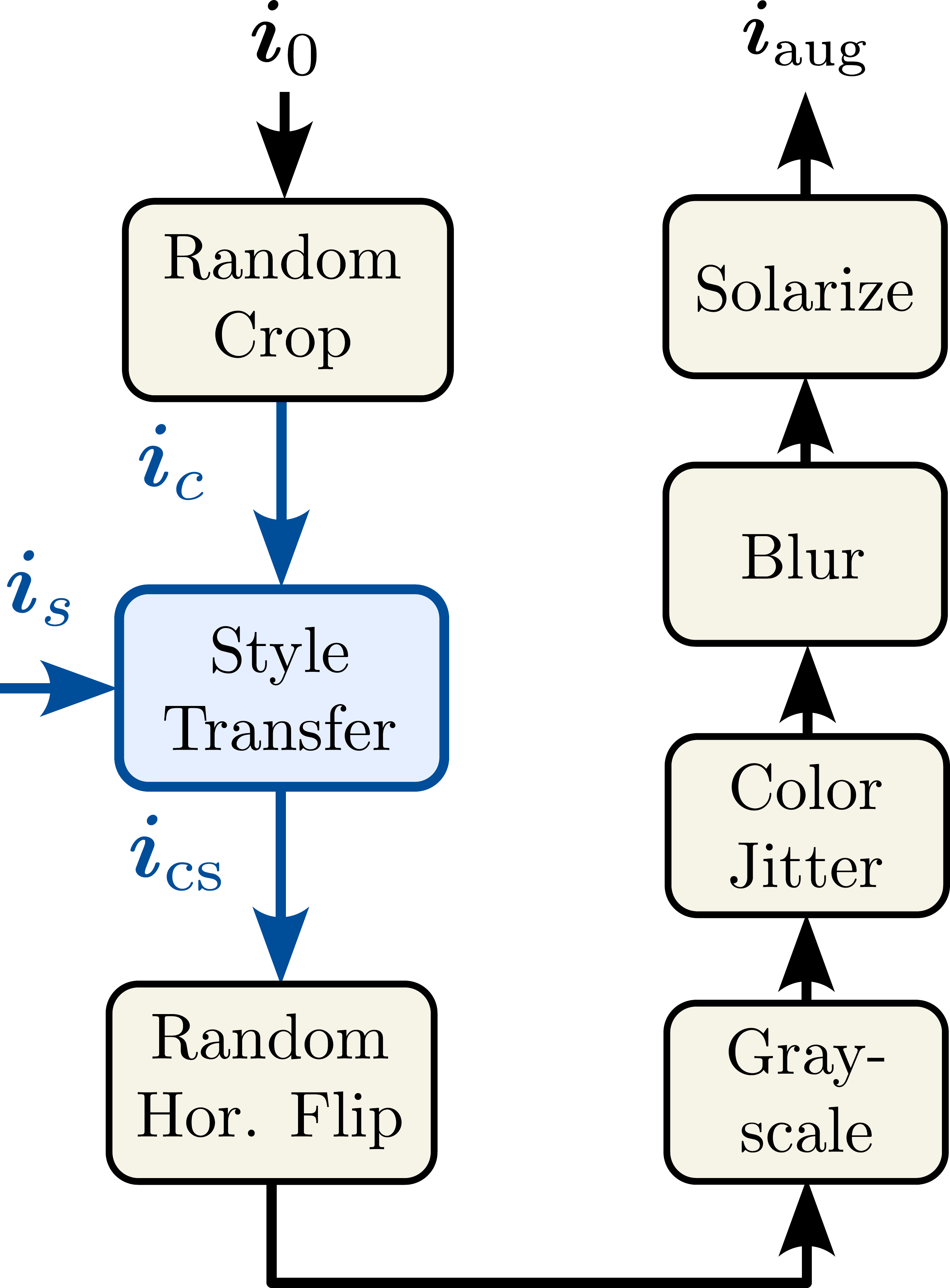}}}
    \subfloat[\textbf{Style Transfer preprocessing}.]{\includegraphics[height=0.2375\linewidth]{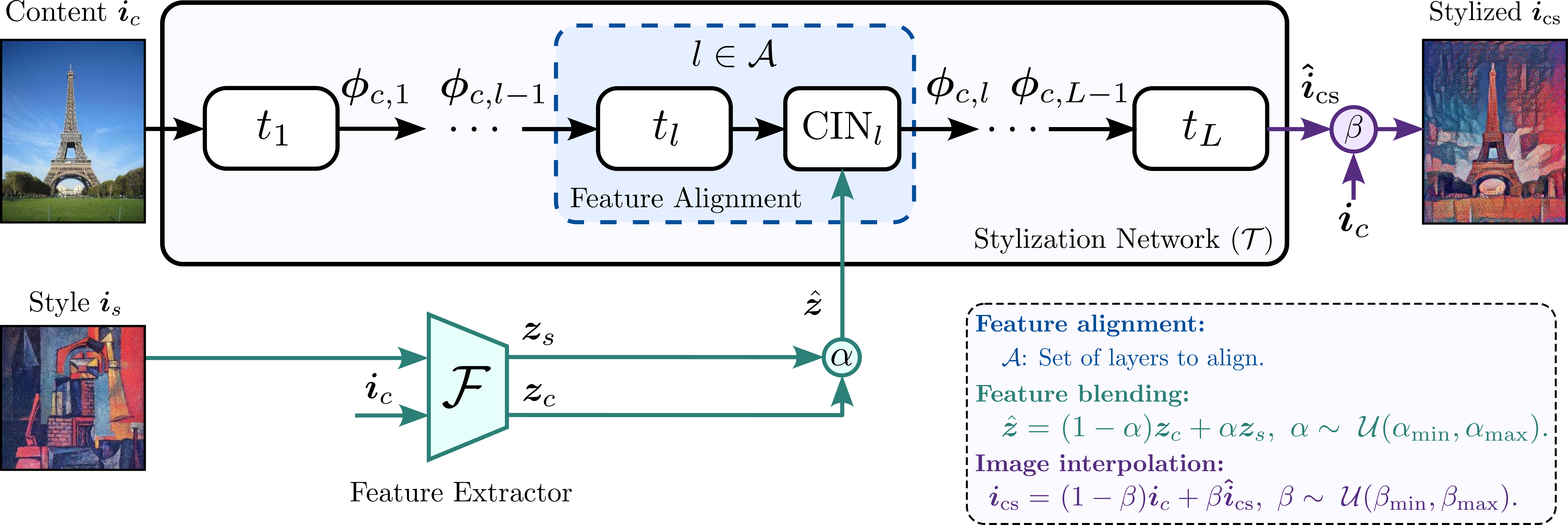}}
    \caption{\textbf{Towards diverse SSL data augmentation via Neural Style Transfer.} We propose \emph{SASSL}, a novel augmentation technique that leverages Style Transfer to create pretraining views that are semantically aware, focusing solely on modifying the image's appearance while preserving its content. SASSL combines the image's content with the texture of an external reference style, generating augmented views that better retain the image's semantic meaning. By incorporating Style Transfer into traditional SSL augmentation pipelines and controlling the stylization strength through gradual blending of style features and pixel values, SASSL promotes stronger representations compared to well-established SSL methods.}
    \label{fig:data_augmentation_pipeline}
    \vspace{-0.3cm}
\end{figure*}

\section{Related Work}
\label{sec:related}
\subsection{Data Augmentation in SSL}
Typical data augmentation methods applied to vision tasks include image cropping and resizing, flipping, rotation, color augmentation, noise addition, and solarization. Examples of methods using these are MoCo \citep{he_2020_momentum}, SimCLR \citep{chen_2020_simple}, BYOL \citep{grill_2020_bootstrap}, and SimSiam \citep{chen_2021_exploring}, among others. Other work \citep{caron_2020_unsupervised} shows how generating additional augmentations using this strategy can improve performance relative to two view approaches, though this strategy decreases throughput and requires additional compute. Other research explore how to select augmentation hyperparameters to learn more robust and general features \citep{wagner_2022_importance, reed_2021_selfaugment, tian_2020_makes}. In contrast, SASSL proposes a new preprocessing technique that can be incorporated in existing augmentation pipelines, boosting performance without additional hyperparameter tuning or changes in pretraining throughput.

Previous SSL work has explored semantic-aware augmentation approaches. \citet{purushwalkam_2020_demystifying} leverage natural video transformations occuring in videos as an alternative to learning from object-centric datasets. \citet{lee_2021_improving} introduces an auxiliary loss to capture the difference between augmented views, leading to better performance on tasks where semantic information is lost due to aggressive augmentation. \citet{bai_2022_rsa} propose an alternative augmentation pipeline to prevent loss of semantics by gradually increasing the strength of augmentations. While these methods modify the pretraining loss and require keeping track of augmentation hyperparameters, SASSL integrates seamlessly into existing pipelines without additional loss terms or auxiliary data. Our method complements default data augmentation pipelines with a content-preserving transformation to obtain stronger image representations.

\subsection{Neural Style Transfer}
Recent image generation and Style Transfer algorithms \citep{liu_2021_adaattn,heitz_2021_sliced,jing_2020_dynamic,yoo_2019_photorealistic,risser_2017_stable,gatys_2017_controlling,chen_2021_artistic} use CNNs to measure the texture similarity between two images in terms of the distance between their neural activations. Feature maps of a pretrained classifier such as VGG-19 \citep{simonyan_2014_very} are extracted and low-order moments of their distribution \citep{kessy_2018_optimal,huang_2017_arbitrary,sheng_2018_avatar} are used as texture descriptors. By matching such feature statistics, these techniques have shown promising results transferring texture between arbitrary images, improving over classic texture synthesis approaches \citep{portilla_2000_parametric, zhu_2000_exploring, heeger_1995_pyramid}.

A large body of work focuses on \emph{artistic} applications, reproducing an artwork style over a scene of interest. These methods adopt either (i) an \emph{iterative optimization} approach \citep{risser_2017_stable,gatys_2016_image,gatys_2017_controlling,li_2017_demystifying}, where an initial guess is gradually transformed to depict a style of interest or (ii) an \emph{autoencoding} approach \citep{liu_2021_adaattn,li_2017_universal,chen_2021_artistic,wang_2020_diversified}, where one or more CNN image generators are trained to impose a target texture in a single forward pass. While selecting an approach implies a trade-off between synthesis quality and computational cost, in both cases the generated stylization shows an unnatural appearance, \ie, it often lacks the qualities of a real-world scene.

In the context of data augmentation for supervised learning, \citet{geirhos_2018_imagenet} and \citet{zheng_2019_stada} addressed texture bias and generated training samples via pre-stylized datasets or stylizing using a small collection of style images. \citet{hong_2021_stylemix} explored Style Transfer to improve robustness against adversarial attacks. \citet{jackson_2019_style} incorporated Style Transfer as a transformation in the augmentation pipeline. While their approach of randomly mixing content and style representations yield promising results, it neglects potential distortions introduced by the Style Transfer network bottleneck. SASSL, in contrast, integrates Style Transfer in a self-supervised setting. Our approach generates diverse augmentations via either pre-computed style representations from external datasets or in-batch stylization with training samples as style references. Importantly, SASSL preserves semantic information through pixel interpolation and feature blending, mitigating the loss of details inherent in Style Transfer networks.

\section{Preliminaries}
\label{sec:prelim}
\subsection{Self-Supervised Learning}
Traditional SSL methods learn compressed representations by maximizing the agreement between differently augmented views of the same data example in a latent space. They do this following the template originally proposed by SimCLR \citep{chen_2020_simple}. In this setup the input is split into multiple views using data augmentations, encoded into a representation, and then further projected into an embedding over which the loss is computed. There are many potential augmentations that can be used, including (but not limited to) random cropping, flipping, color jitter, blurring, and solarization. By maximizing the similarity of augmented training samples, the network learns to create robust representations that separate meaningful semantic content from simple distortions that could occur in the real world, and which should not affect the semantic content of an image. 

Given a batch of $N$ input images $\{\vi_{k}\}_{k=1}^{N}$, $2N$ augmented samples are generated by applying distinct transformations to each image. These transformations correspond to the same data augmentation pipeline. Let $R$ correspond to all possible augmentations. Then, positive pairs correspond to augmented views of the same input sample, and negative pairs correspond to views coming from different input images. Based on this, the $2N$ augmented samples $\{\tilde{\vi}_{l}\}_{l=1}^{2N}$ can be organized so that indices $l=2k-1$ and $l=2k$ correspond to views of the $k$-th input sample
\begin{align}
    \tilde{\vi}_{2k-1}=\ r(\vi_{k}),\quad \tilde{\vi}_{2k}=\ \hat{r}(\vi_{k}),\quad \hat{r},r \sim R
\end{align}

Once augmented views are obtained, a representation is computed using an image encoder (typically a CNN model). The representations are then fed to a projection head which further compresses them into a lower-dimensional manifold where different views of the same image are close together and those from different images are far apart. Let $h$ and $g$ be the encoder (e.g. a ResNet-50 backbone) and projection head (e.g. an MLP layer), respectively. Then, embeddings are obtained for each augmented sample as $\vz_{l}=g\circ h(\tilde{\vi}_{l})$.

SimCLR uses the normalized temperature-scaled cross entropy loss (NT-Xent) to learn how to identify positive pairs of augmented samples. First, the cosine similarity of every pair of embeddings is computed
\begin{align}
    s_{m,n}=&\ \frac{\langle \vz_{m},\vz_{n}\rangle}{\|\vz_{m}\|\|\vz_{n}\|}
\end{align}
The model is then trained using a contrastive loss by comparing the embeddings of positives, forcing them to be similar to each other. Since the loss is normalized, it naturally forces the representations of views from two different images (negatives) to be distant from each other.
\begin{align}
    \mathcal{L}=&\ \frac{1}{2N}\sum_{k=1}^{N}\big[\ell(2k-1,2k)+\ell(2k,2k-1)\big],\quad
    \ell(m,n)=\ -\log\bigg(\frac{\exp(s_{m,n}/\tau)}{\sum_{l=1}^{2N}\mathbbm{1}_{m\neq n}\exp(s_{m,l}/\tau)}\bigg)
\end{align}
where $\tau \in \mathbb{R}_{++}$ is the temperature factor and $\mathbbm{1}$ the indicator function. While SimCLR is a simple framework, it pushed the state-of-the-art significantly on a wide range of downstream tasks including image classification, object detection, and semantic segmentation.

Follow up works to SimCLR such as MoCo \citep{chen_2020_improved, chen_2021_empirical}, BYOL \citep{grill_2020_bootstrap} and SimSiam \citep{chen_2021_exploring}, among others \citep{caron_2020_unsupervised, caron_2021_emerging, assran_2022_masked, zbontar2021barlow, bardes_2021_vicreg}, have largely maintained this template, but have proposed modifications to this setup (e.g. new losses, architectures, or augmentation strategies) which attempt to further improve the downstream task performance.

\subsection{Neural Style Transfer}
Style Transfer techniques combine the semantics (\emph{content}) of an image with the visual characteristics (\emph{style}) of another image. These assume that the statistics of shallower layers of a trained CNN encode style, while deeper layers encode content. Seminal techniques are based on an optimization-based approach, passing a pair of content and style images to a CNN encoder and optimizing over a randomly initialized image to produce activations with similar statistics to the style image at shallower layers and similar activations to the content image at deeper ones \citep{gatys_2015_neural}. This way, a \emph{stylized} image is generated, comprising the semantic and texture attributes of interest.

While optimization-based methods generate a diverse stylization due to a random image initialization, autoencoding methods utilize an image decoder to efficiently stylize arbitrary image pairs on a single forward pass. In what follows, we introduce the autoencoding Style Transfer technique adopted by our proposed method due on its generalization and efficiency properties. For an in-depth survey of Style Transfer methods, refer to \citet{jing_2019_neural}.

{\noindent \bf Fast Style Transfer.} \citet{dumoulin_2017_learned} proposed an arbitrary Style Transfer method with remarkable generalization properties. Their algorithm, \emph{Fast Style Transfer}, accurately represents unseen artistic styles by training a model to predict first and second moments of latent image representations at multiple scales. Such moments are used as arguments of a special form of instance normalization, denominated \emph{conditional instance normalization} (CIN), to impose style over arbitrary input images.

Given a content image $\vi_{c}\in \mathbb{R}^{C\times H_{c}\times W_{c}}$ and a style image $\vi_{s}\in \mathbb{R}^{C\times H_{s}\times W_{s}}$, Fast Style Transfer produces a stylized image $\vi_{\text{cs}}$ that corresponds to
\begin{align}
    \vi_{\text{cs}}=&\  \mathcal{T}(\vi_{c},\vz_{s})\in \mathbb{R}^{C\times H_{c}\times W_{c}}
\end{align}

where $\mathcal{T}$ is a stylization network and $\vz_{s}=\mathcal{F}(\vi_{s})\in \mathbb{R}^{D}$ is an embedding extracted from the style image via a feature extractor $\mathcal{F}$, e.g., InceptionV3 \citep{szegedy_2016_rethinking}.

We assume $\vz_{s}$ to be a contracted embedding of the style image ($D \ll CH_{s}W_{s}$). The stylization network $\mathcal{T}$ is comprised by $L$ blocks $\{t_{l}\}_{l=1}^{L}$. $\mathcal{T}$ extracts high-level features from the content image, aligns them to the style embedding $\vz_{s}$ and maps the resulting features to the pixel domain. The style of $\vi_{s}$ encapsulated in $\vz_{s}$ is transferred to the content image using CIN. This is applied to a particular set of layers to impose the target texture and color scheme by aligning feature maps at different scales. \begin{wrapfigure}[24]{r}{0.495\textwidth}
\includegraphics[width=0.495\textwidth]{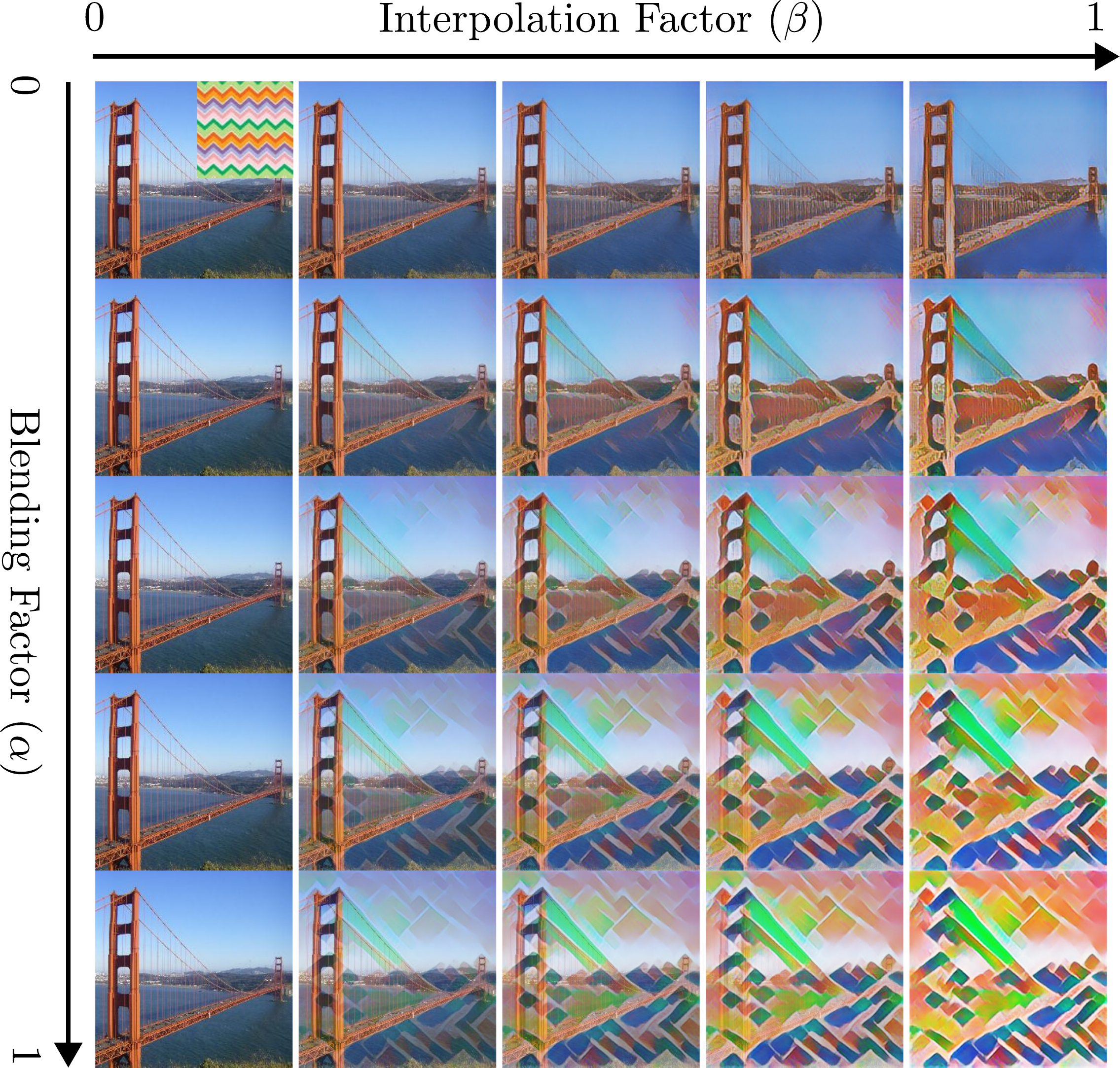}
\captionof{figure}{\label{fig:blend_interpolate}\textbf{Feature blending and image interpolation}. A fine-grained control over the final stylized image is obtained via interpolation and blending factors $\alpha$ and $\beta$ that operate in the feature and pixel domains. This prevents augmented views from losing semantic information due to strong transformations.}
\end{wrapfigure}
\vspace{-0.45 cm}

We define the set of layers where CIN is applied as $\mathcal{A}$. The normalization imposed via CIN consists of an extended form of instance normalization where the target mean and standard deviation are extracted from a style representation $\vz$.

Given an input $\vi\in \mathbb{R}^{C\times H\times W}$ and a style representation $\vz\in \mathbb{R}^{D}$, CIN is defined as
\begin{align}
    \hat{\vi}=&\ \text{CIN}(\vi,\vz)\in \mathbb{R}^{C\times H\times W}\\
    \hat{\vi}^{(k)}=&\ \gamma^{(k)}(\vz)\bigg(\frac{\vi^{(k)}-\mathbb{E}[\vi^{(k)}]}{\sigma(\vi^{(k)})}\bigg)+\lambda^{(k)}(\vz)
\end{align}
where $\vi^{(k)},\ k\in \{1, \dots, C\}$ corresponds to the $k$-th input channel, and the sample mean $\mathbb{E}[\vi^{(k)}]$ and standard deviation $\sigma(\vi^{(k)})$ are computed along its spatial support. Here, $\gamma^{(k)}, \lambda^{(k)}: \mathbb{R}^{D}\mapsto \mathbb{R}$ are trainable functions that predict scaling and offset values from the latent representation $\vz$ for the $k$-th input channel. The layers in $\mathcal{T}$ are characterized by
\begin{align}
    \bm{\phi}_{c,l}=&\ \begin{cases}
        \text{CIN}_{l}\big(t_{l}(\bm{\phi}_{c,l-1}), \vz_{s}\big), & l\in \mathcal{A} \\
        t_{l}(\bm{\phi}_{c,l-1}), & l \notin \mathcal{A}
    \end{cases}
\end{align}

where the input of the stylization network corresponds to $\bm{\phi}_{c,0}=\vi_{c}$. The subscript $l$ in the CIN operation indicates that each layer has its own $\gamma$ and $\lambda$ functions to normalize features independently.

Our method uses the Fast Style Transfer algorithm. Given its generalization properties and low-dimensional style representations, it is a good match for our framework, where style representations from multiple domains must be efficiently extracted, manipulated and transferred.

\section{Proposed Method: SASSL}
\label{sec:proposed}
We provide a detailed description of our SSL augmentation technique. First, we break down SASSL's key components and hyperparameters. Then, we tackle the problem of making the augmented images more diverse by utilizing style references from different domains in an efficient manner.

\begin{minipage}[t]{0.48\textwidth}
      \begin{algorithm}[H]
        \small
        \caption{Style transfer augmentation block}
        \SetCustomAlgoRuledWidth{0.45\textwidth}
        \label{alg:style_transfer}
        {\bfseries Input:} $\vi_{c},\vi_{s},\mathcal{F},\mathcal{T},\alpha_{\text{min}},\alpha_{\text{max}},\beta_{\text{min}},\beta_{\text{max}}$
        
        {\bfseries Output:} $\vi_{\text{cs}}$
        \BlankLine\BlankLine
        $\vz_{c}\gets \mathcal{F}(\vi_{c})$ \tcp*[r]{Style representation of content image}
        
        $\vz_{s}\gets \mathcal{F}(\vi_{s})$ \tcp*[r]{Style representation of style image}
        \BlankLine\BlankLine\BlankLine\BlankLine
        $\alpha\sim\mathcal{U}(\alpha_{\text{min}},\alpha_{\text{max}})$ \tcp*[r]{Blending factor}
        $\hat{\vz}\gets (1-\alpha)\vz_{c}+\alpha \vz_{s}$ \tcp*{Feature blending}
        \BlankLine\BlankLine\BlankLine\BlankLine
        $\hat{\vi}_{\text{cs}}\gets\ \mathcal{T}(\vi_{c},\hat{\vz})$\;
        $\beta\sim\mathcal{U}(\beta_{\text{min}},\beta_{\text{max}})$ \tcp*{Interpolation factor}
        $\vi_{\text{cs}}\gets (1-\beta)\vi_{c}+\beta \hat{\vi}_{\text{cs}}$ \tcp*{Stylized image}
      \end{algorithm}
\end{minipage}
\hspace{0.025\textwidth}
\begin{minipage}[t]{0.48\textwidth}
\vspace{-0.15 cm}
\centering
\hspace{-0.075\baselineskip}\colorbox{white}{\begin{minipage}[t]{0.175\columnwidth}\vspace{-0.5\baselineskip}
\centering \textbf{\scalebox{0.625}{\makecell{Content\\Image}}}
\end{minipage}}\colorbox{white}{\begin{minipage}[t]{0.175\columnwidth}\vspace{-0.5\baselineskip}
\centering \textbf{\scalebox{0.625}{\makecell{Style\\ (In-batch)}}}
\end{minipage}}\colorbox{white}{\begin{minipage}[t]{0.175\columnwidth}\vspace{-0.5\baselineskip}
\centering \textbf{\scalebox{0.625}{\makecell{Style\\ (External)}}}
\end{minipage}}\colorbox{white}{\begin{minipage}[t]{0.175\columnwidth}\vspace{-0.5\baselineskip}
\centering\textbf{\scalebox{0.625}{\makecell{Stylized\\ (In-batch)}}}
\end{minipage}}\colorbox{white}{\begin{minipage}[t]{0.175\columnwidth}\vspace{-0.5\baselineskip}
\centering\textbf{\scalebox{0.625}{\makecell{Stylized\\ (External)}}}
\end{minipage}}
    \vspace{-0.1 cm}
    \centering
    \includegraphics[width=\columnwidth]{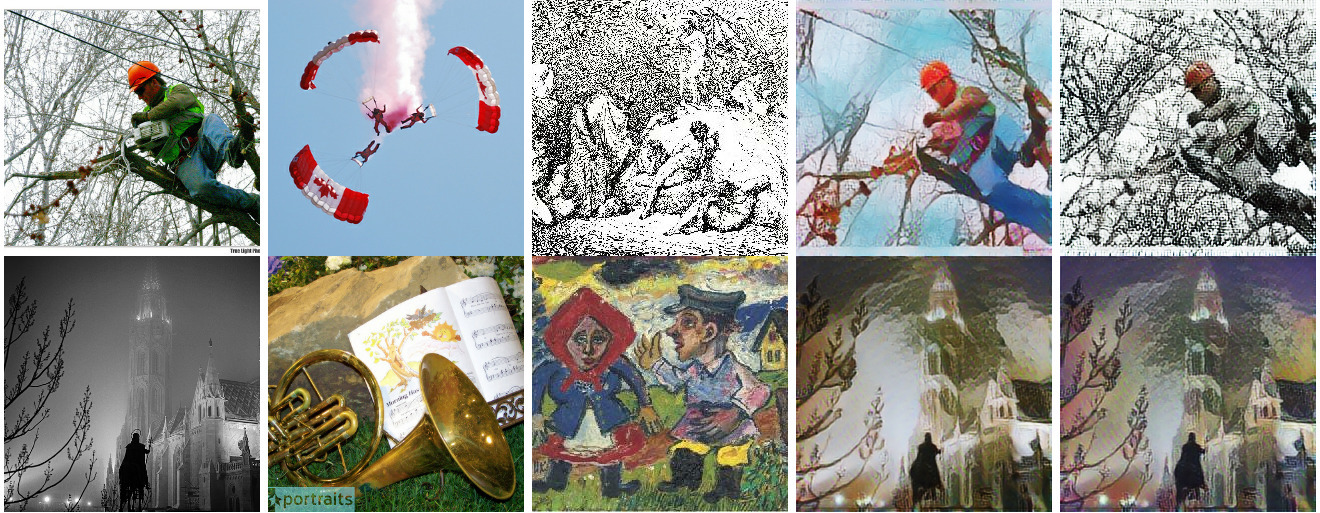}
    \vspace{-0.4 cm}
    \captionof{figure}{\label{fig:stylization}\textbf{Style Transfer examples}. Views generated using style references from the same domain \textit{(in-batch)} as well as other domains \textit{(external)}. Stylization obtained using a blending factor $\alpha=0.5$.}
\end{minipage}
\vspace{0.15 cm}

{\bf Style Transfer as data preprocessing.} We incorporate Style Transfer to the default preprocessing pipeline of SSL methods. It is worth noting that SASSL is not specific to a particular SSL approach, and can be readily applied with different methods. Figure \ref{fig:data_augmentation_pipeline} shows an example of our augmentation pipeline, where Style Transfer is applied after random cropping. A raw input image $\vi_{0}$ is cropped, producing a view that is taken as the content image $\vi_{c}$. Given an arbitrary style image $\vi_{s}$ (we discuss the choice of $\vi_{s}$ below), the Style Transfer block generates a stylized image $\vi_{\text{cs}}$ by imposing the texture of $\vi_{s}$ over $\vi_{c}$. Finally, the stylized image $\vi_{\text{cs}}$ is passed to the remaining augmentation blocks to produce an augmented sample $\vi_{\text{aug}}$.

As discussed in recent work on SSL augmentation \citep{han_2022_you, chen_2020_simple}, adding a strong transformation to a self-supervised method tends to degrade performance. For this reason, it is crucial to control the amount of stylization imposed in the augmentation stage. We do so by introducing three hyperparameters: probability $p\in [0,1]$, which dictates whether an image is stylized or not, a blending factor $\alpha \in [0,1]$ to combine content $\vz_{c}$ and style $\vz_{s}$ representations, and an interpolation factor $\beta \in [0,1]$ to combine content $\vi_{c}$ and stylized $\hat{\vi}_{\text{cs}}$ images.

Given style representations extracted from content and style images $\vz_{c}=\mathcal{F}(\vi_{c})$ and $\vz_{s}=\mathcal{F}(\vi_{s})$, respectively, we obtain an intermediate stylized image $\hat{\vi}_{\text{cs}}$ by applying a convex combination based on blending factor $\alpha$.
\begin{align}
    \hat{\vi}_{\text{cs}}=&\ \mathcal{T}(\vi_{c},\hat{\vz}),\quad
    \hat{\vz}=\ (1-\alpha)\vz_{c}+\alpha \vz_{s}
\end{align}

Then, the final stylization output is obtained via a convex combination between the intermediate stylized image $\hat{\vi}_{cs}$ and the content image $\vi_{c}$ based on interpolation factor $\beta$.
\begin{align}
    \vi_{\text{cs}}=&\ (1-\beta)\vi_{c}+\beta \hat{\vi}_{\text{\text{cs}}}
\end{align}

Algorithm \ref{alg:style_transfer} describes our proposed Style Transfer data augmentation block. Figure \ref{fig:blend_interpolate} illustrates the effect of the feature blending and image interpolation operations, showcasing their importance to control the stylization effect without degrading the semantic attributes.

SASSL operates over minibatches, allowing efficient data pre-processing. Let $\mI_{c}\in \mathbb{R}^{B\times C\times H_{c}\times W_{c}}$ and $\mI_{s}\in \mathbb{R}^{B\times C\times H_{s}\times W_{s}}$ be content and style minibatches, respectively, comprised by $B$ images $\mI_{c}^{(b)}$ and $\mI_{s}^{(b)}, b\in \{1, \dots, B\}$. Then, the stylized minibatch $\mI_{\text{cs}}\in \mathbb{R}^{B\times C\times H_{c}\times W_{c}}$ is generated by applying Style Transfer between a sample from the content batch and a sample from the style batch, given an arbitrary selection criterion. We propose two alternatives for selecting style images to balance between augmentation diversity and efficiency.

{\bf Diversifying style references.} In contrast to traditional data augmentation, Style Transfer can leverage a second dataset to extract style references. This opens the possibility of selecting style images from different domains, diversifying the transformations applied to the pretraining dataset. SASSL relies on two approaches for sampling style references: \emph{external} and \emph{in-batch} stylization.

\emph{External} stylization consists on pre-computing representations of an arbitrary style dataset and sampling from them during pretraining. This allows controlling the styles to impose on the augmented views while reducing the computational overhead of Style Transfer. Under this configuration, the Style Transfer block receives a content minibatch along with a minibatch of pre-computed style representations extracted from an arbitrary style dataset, and generates a minibatch of stylized images.

On the other hand, \emph{in-batch} stylization uses the styles depicted in the content dataset itself by using other images of the content minibatch as style references. This is of particular interest for large-scale pretraining datasets covering multiple image categories and thus textures (e.g. ImageNet). So, enabling the use of a single dataset for both pretraining and stylization is a valid alternative.

Following this, samples from the same minibatch can be used as style references by associating pairs of images in a circular fashion. More precisely, a style minibatch $\mI_{s}\in \mathbb{R}^{B\times C\times H_{c}\times W_{c}}$ is generated by applying a circular shift on the content minibatch indices
\begin{align}
    \mI_{s}^{(b)}=&\ \mI_{c}^{((b-b_{0})\ \text{mod}\ B)}
\end{align}
where $\text{mod}$ denotes the modulo operation and $b_{0}$ is an arbitrary offset. Figure \ref{fig:stylization} shows ImageNet samples stylized using pre-computed style representations from the Painter by Numbers dataset \citep{painter_by_numbers} via \emph{external} stylization, as well as using other ImageNet samples taken from the content minibatch via \emph{in-batch} stylization.
\begin{wrapfigure}{r}{0.5\textwidth}
\vspace{1.35 cm}
\centering
\small
\rowcolors{2}{bg_blue}{}
\setlength{\tabcolsep}{7.5pt}
\centering
\captionof{table}{\textbf{SASSL $+$ MoCo v2 downstream classification performance on ImageNet}. Linear probing accuracy ($\%$) of a ResNet-50 backbone pretrained using SASSL $+$ MoCo v2. Mean accuracy reported over five random trials.}
\vspace{-0.2 cm}
\resizebox{0.5\textwidth}{!}{
\begin{tabular}{cccccc}
\toprule
Method & Top-1 Acc. & Top-5 Acc.\\ 
\midrule
    \makecell{MoCo v2 (Default)} & $72.55$ & $91.19$\\
    \makecell{SASSL $+$ MoCo v2 \textbf{(Ours)}} &   $\bf 74.64$ & $\bf 91.68$\\
\bottomrule
\end{tabular}}
\label{tab:downstream_performance}
\vspace{-0.25 cm}
\end{wrapfigure}

\section{Experiments}
\label{sec:experiments}
\subsection{Downstream Task Performance}
\label{sec:downstream_task_performance}
We evaluate the downstream ImageNet classification accuracy of SSL models pretrained via SASSL on the MoCo framework. We compare a MoCo v2 model pretrained with our data augmentation vs. a MoCo v2 baseline with default augmentation \citep{chen_2020_improved}. Note that MoCo v2 and SimCLR use the same loss, architecture, and augmentations (they differ by MoCo's momentum encoding).

{\noindent\bf Pretraining settings.} Our pretraining setup is similar to the canonical SSL setup used to pretrain SimCLR and BYOL. We use the same loss, architecture, optimizer, and learning rate schedule as MoCo v2 for fair comparison. We pretrain a ResNet-50 encoder on ImageNet for $1,000$ epochs via SASSL. To measure downstream accuracy, we add a linear classification head on top of the pretrained backbone and train in a supervised fashion on ImageNet.

SASSL pretraining applies Style Transfer only to the left view (no changes in augmentation are applied to the right view). It is applied with a probability $p=0.8$ using blending and interpolation factors drawn from a uniform distribution $\alpha, \beta \sim \mathcal{U}(0.1, 0.3)$. We found that this modest stylization best complimented the existing augmentations, avoiding overly-strong transformations that can hinder performance \citep{han_2022_you, chen_2020_simple}.

{\noindent\bf Results.} Table \ref{tab:downstream_performance} compares the downstream classification accuracy obtained by our SASSL augmentation approach on MoCo v2 using \emph{external} stylization from the Painter by Numbers dataset. Results indicate our proposed augmentation improves downstream task performance by $2.09\%$ top-1 accuracy. This highlights the value of Style Transfer augmentation in self-supervised training, where downstream task performance significantly boosts by incorporating transformations that decouple content and style. We also report results with \emph{in-batch} stylization in \cref{sec:ablation_study}.

\subsection{Transfer Learning Performance}
\label{sec:transfer_learning_performance}
To better understand the robustness and generalization of image representations learned using our proposed data augmentation approach, we evaluate its transfer learning performance across various tasks. By incorporating Style Transfer, we hypothesize that the learned representations become invariant to changes in appearance such as color and texture. This forces the feature extraction process to rely exclusively on semantic attributes. As a result, the learned representations may become more robust to domain shifts, improving downstream task performance across datasets. We empirically show this property by evaluating the transfer learning performance of image representations trained using SASSL under linear probing and fine-tuning scenarios.

{\noindent\bf Downstream settings.} We compare the transfer learning accuracy of ResNet-50 pretrained via MoCo v2 using SASSL against a MoCo v2 baseline with default data augmentation. The evaluated models are pretrained on ImageNet and transferred to eleven target datasets: ImageNet-1\% subset \citep{chen_2020_simple}, iNaturalist `21 (iNat21) \citep{inaturalist21}, Diabetic Retinopathy Detection (Retinopathy) \citep{kaggle_diabetic_retinopathy}, Describable Textures Dataset (DTD) \citep{cimpoi_2014_describing}, Food101 \citep{bossard14}, CIFAR10/100 \citep{Krizhevsky09learningmultiple}, SUN397 \citep{Xiao:2010}, Cars \citep{KrauseStarkDengFei-Fei_3DRR2013}, Caltech-101 \citep{FeiFei2004LearningGV}, and Flowers \citep{Nilsback08}.

To have a clear idea of the effect of the style dataset in SASSL's pipeline, we pretrain five ResNet-50 backbones, each using a different style. We use ImageNet, iNat21, Retinopathy, DTD, and Painter by Numbers (PBN) as style datasets. More precisely, we transfer five models, each pretrained on a different style, to each of eleven target datasets. We also include ImageNet as target dataset to compare the effect of different styles on downstream task performance. This leads to $60$ transfer learning scenarios used to better understand the effect of various styles on different image domains.

Transfer learning is evaluated in terms of top-1 classification accuracy on linear probing and fine-tuning. All models were pretrained as described in \cref{sec:downstream_task_performance}. We report mean accuracy across five trials. Please refer to \cref{sec:supp_downstream_settings} for full linear probing and fine-tuning training and testing settings.

\begin{table*}
\centering
\setlength{\tabcolsep}{2.0pt}
\small
\caption{\textbf{Transfer learning performance.} Downstream top-1 classification accuracy ($\%$) of SASSL $+$ MoCo v2 pretrained on ImageNet. Our data augmentation method generates specialized image representations that improve transfer learning performance, as shown in linear probing and fine-tuning scenarios. Mean accuracy reported over five random trials.}
\resizebox{\linewidth}{!}{%
\begin{tabular}{cccccccccccccc}
\toprule
& & \multicolumn{12}{c}{\textbf{Target Dataset}}\\
& & ImageNet & ImageNet-1\% & iNat21 & Retinopathy & DTD & Food101 & CIFAR10 & CIFAR100 & SUN397 & Cars & Caltech-101 & Flowers\\
\midrule
 & & \multicolumn{12}{c}{\emph{Linear Probing}}\\
\cmidrule{2-14}
  & None (Default)  & $72.55$ & $53.23$ & $41.33$ & $\bf75.88$ & $72.68$ & $ 73.82$ & $ 89.94$ & $ 71.93$ & $ 69.96$ & $ 53.15$ & $ 88.19$ & $ 93.39$\\
  \rowcolor{bg_blue}\cellcolor{white} & \textcolor{black}{ImageNet (\textbf{Ours})} & $74.07$ & $56.87$ & $45.01$ & $75.75$ & $73.69$ & $74.43$ & $ 90.93$ & $73.26$ & $69.67$ & $\bm{64.87}$ & $89.3$ & $95.27$\\
  \rowcolor{bg_blue}\cellcolor{white} & \textcolor{black}{iNat21 (\textbf{Ours})} & $74.28$ & $56.76$ & $44.70$ & $75.75$ & $72.75$ & $ 74.3$ & $\bm{91.04}$ & $ 73.29$ & $\bm{70.07}$ & $ 63.96$ & $\bm{89.89}$ & $94.7$\\
  \rowcolor{bg_blue}\cellcolor{white} & \textcolor{black}{Retinopathy (\textbf{Ours})} & $74.02$ & $\bf 56.99$ & $44.9$ & $75.78$ & $73.73 $ & $74.53$ & $90.8$ & $73.3$ & $69.63$ & $64.06$ & $89.17$ & $ 94.94$\\
  \rowcolor{bg_blue}\cellcolor{white} & \textcolor{black}{DTD (\textbf{Ours})} & $74.32$ & $56.77$ & $\bf45.08$ & $75.76$ & $\bf74.41$ &$\bm{74.88}$ & $\bf{91.04}$ & $\bm{73.41}$ & $69.71$ & $64.58$ & $89.3$ & $95.24$\\
  \rowcolor{bg_blue}\cellcolor{white} & \textcolor{black}{PBN (\textbf{Ours})} & $\bf74.64$ & $56.9$ & $ 45.02$ & $75.79$ & $72.77$ & $74.37$ & $90.85$ & $73.38$ & $69.69$ & $64.12$ & $89.59$ & $\bm{95.45}$\\
\cmidrule{2-14}
& & \multicolumn{12}{c}{\emph{Fine-tuning}}\\
\cmidrule{2-14}
& None (Default)  & $74.89$ & $51.61$ & $77.92$ & $78.89$ & $71.54$ & $87.25$ & $96.91$ & $83.4$ & $74.25$ & $83.63$ & $89.27$ & $95.75$\\
  \rowcolor{bg_blue}\cellcolor{white} & \textcolor{black}{ImageNet (\textbf{Ours})} & $75.52$ & $51.74$ & $79.21$ & $79.64$ & $\bf 72.31$ & $ 87.48$ & $ 97.0$ & $83.21$ & $73.89$ & $\bm{90.33}$ & $88.26$ & $\bm{96.6}$\\
  \rowcolor{bg_blue}\cellcolor{white} & \textcolor{black}{iNat21 (\textbf{Ours})} & $\bf 75.58$ & $\bf 51.86$ & $79.19$ & $79.6$ & $71.35$ & $87.4$ & $\bm{97.05}$ & $83.29$ & $74.05$ & $90.04$ & $88.55$ & $95.76$\\
  \rowcolor{bg_blue}\cellcolor{white} & \textcolor{black}{Retinopathy (\textbf{Ours})} & $75.52$ & $51.76$ & $79.23$ & $79.63$ & $72.07$ & $87.39$ & $96.97$ & $\bm{83.68}$ & $\bm{74.26}$ & $89.96$ & $ 88.44$ & $96.34$\\
  \rowcolor{bg_blue}\cellcolor{white} & \textcolor{black}{DTD (\textbf{Ours})} & $75.24$ & $51.73$ & $\bf 79.24$ & $\bf 79.7$ & $70.59$ & $\bm{87.66}$ & $96.77$ & $83.28$ & $74.17$ & $89.59$ & $\bm{89.54}$ & $95.59$\\
  \rowcolor{bg_blue}\cellcolor{white} \multirow{-14}{*}{\rotatebox[origin=c]{90}{\textbf{Style Dataset}}} & \textcolor{black}{PBN (\textbf{Ours})} & $75.05$ & $51.85$ & $ 79.2$ & $79.63$ & $71.35$ & $87.56$ & $96.97$ & $83.36$ & $74.18$ & $89.75$ & $88.97$ & $95.77$\\
\bottomrule
\end{tabular}}
\label{tab:transfer_learning_full}
\end{table*}

{\noindent\bf Results.} Table \ref{tab:transfer_learning_full} shows the top-1 classification accuracy obtained via transfer learning. For linear probing, SASSL significantly improves the average performance on eleven out of twelve target datasets by up to $10\%$ top-1 classification accuracy. For Retinopathy, SASSL obtains on-par linear probing accuracy to the default MoCo v2 model.

For fine-tuning, all models trained via SASSL outperform the baseline. Results show the average top-1 classification accuracy improves by up to $6\%$. This suggests SASSL generalizes across datasets, spanning from textures (DTD) to medical images (Retinopathy). Note that, for a fair comparison, we do not perform hyperparameter tuning.  Interestingly, the relative performance obtained from different style datasets generally differs comparably to the measurement uncertainty, which is shown for these experiments in Section~\ref{sec:supp_additional_mocov2_results} of the supplementary material.  This suggests that the choice of style dataset is secondary in importance, while the main benefit comes from the use of SASSL itself.

\begin{wrapfigure}[7]{r}{0.5\textwidth}
\vspace{-1.375 cm}
\setlength{\tabcolsep}{2pt}
\rowcolors{2}{bg_blue}{}
\centering
\captionof{table}{\textbf{SASSL + MoCo v2 Few-shot learning performance}. One and ten-shot top-1 classification accuracy ($\%$) of representations learned via SASSL $+$ MoCo v2. Accuracy reported on a single trial.}
\vspace{-0.1 cm}
\resizebox{0.5\textwidth}{!}{
\begin{tabular}{cccccc}
\toprule
Method & One-shot Acc. & Ten-shot Acc.\\ 
\midrule
    MoCo v2 (Default) & $19.56$ & $45.05$\\
    SASSL $+$ MoCo v2 \textbf{(Ours)} & $\bf 20.55$ & $\bf46.73$\\
\bottomrule
\end{tabular}}
\label{tab:fewshot}
\end{wrapfigure}
\vspace{0.25 cm}

\subsection{Few-shot Learning Performance}
To further demonstrate the representation learning capabilities of the data augmentation imposed via SASSL, we conduct experiments on \textit{few-shot classification}. We compare our ResNet-50 backbone pretrained via SASSL $+$ MoCo v2 against a MoCo v2 baseline in the context of one and ten-shot learning on ImageNet.

Table \ref{tab:fewshot} shows the few-shot classification accuracy. Results reveal that SASSL boosts few-shot classification top-1 accuracy by over $1\%$ in both one and ten-shot learning. This aligns with our previous experiments, suggesting that SASSL promotes more general image representations.

\subsection{Additional Downstream Performance Evaluation}
\label{sec:additional_downstream_performance}
{\noindent\bf Performance on other SSL methods.} To assess SASSL's broader impact, we evaluate its effectiveness on two other SSL methods, SimCLR and BYOL. We pretrain ResNet-50 backbones with each method, and then use linear probing on ImageNet to compare the quality of their learned representations. For each method, default pretraining and linear probing configurations are used. For SASSL, we employ its recommended hyperparameters ($\alpha, \beta \in [0.1, 0.3]$, $p=0.8$) and PBN as style dataset.

Table \ref{tab:additional_downstream_performance} shows the accuracy attained by SimCLR and BYOL equipped with SASSL. Results show our proposed data augmentation technique boosts top-1 accuracy by approximately $1\%$ in both cases, highlighting its potential across multiple SSL techniques.

\begin{minipage}[t]{0.48\textwidth}
\centering
\small
\rowcolors{2}{bg_blue}{}
\setlength{\tabcolsep}{9.75pt}
\centering
\captionof{table}{\textbf{SASSL downstream performance using alternative SSL methods}. Linear probing accuracy ($\%$) on ImageNet using a ResNet-50 backbone pretrained via SimCLR and BYOL. Accuracy reported on a single trial.}
\resizebox{\columnwidth}{!}{
\begin{tabular}{cccccc}
\toprule
Method & Top-1 Acc. & Top-5 Acc. \\ 
\midrule
    SimCLR (Default) & $68.62$ & $88.7$\\
    SASSL $+$ SimCLR \textbf{(Ours)} &   $\bf 69.58$ & $\bf89.01$\\
\midrule
    BYOL (Default)  & $74.09$ & $91.83$\\
    SASSL $+$ BYOL \textbf{(Ours)} & $\bf 75.13$ & $\bf 92.12$\\
\midrule
    SwAV (Default) & $70.45$ & $89.6$\\
    SASSL $+$ SwAV \textbf{(Ours)} & $\bf 71.3$ & $\bf 90.39$\\
\bottomrule
\end{tabular}}
\label{tab:additional_downstream_performance}
\end{minipage}
\hspace{0.02\textwidth}
\begin{minipage}[t]{0.48\textwidth}
\centering
\small
\rowcolors{2}{bg_blue}{}
\setlength{\tabcolsep}{1.5pt}
\centering
\captionof{table}{\textbf{SASSL $+$ MoCo downstream performance using alternative backbones}. Linear probing accuracy ($\%$) on ImageNet using ResNet-50 (x4) and ViT-B/16 representation models. Accuracy reported on a single trial.}
\resizebox{\linewidth}{!}{
\begin{tabular}{cccccc}
\toprule
Backbone & Method & Top-1 Acc. & Top-5 Acc.\\ 
\midrule
    ResNet-50 & MoCo v2 (Default) & $77.2$ & $93.32$\\
    \cellcolor{white} x4 ($375$M) & \makecell{SASSL $+$ MoCo v2 \textbf{(Ours)}} &  $\bf 78.21$ & $\bf 93.98$\\
\midrule
    ViT-B/16 & MoCo v3 (Default) & $75.01$ & $92.43$\\
    \cellcolor{white} ($86$M) & SASSL $+$ MoCo v3 \textbf{(Ours)} & $\bf 75.51$ & $\bf 92.56$\\
\bottomrule
\end{tabular}}
\label{tab:downstream_performance_arch}
\vspace{1 cm}
\end{minipage}

{\noindent\bf Performance on other representation models.}
We explore SASSL's performance on models with varying complexity and architecture. For complexity, we employ ResNet-50 (x4), a scaled-up version of the previously evaluated ResNet-50 (from 24 to 375 million parameters). This allows us to probe how SASSL scales with increased model size. In terms of architecture, we employ ViT-B/16, a Transformer-based backbone with 86 million parameters and a distinct design compared to previous CNN models.

We pretrain and linearly probe a ResNet-50 (x4) representation model on ImageNet via MoCo v2. Pretraining and downstream settings follow our default configuration, as documented in the \cref{sec:supp_additional_experimental_details,sec:supp_downstream_settings}. Similarly, we pretrain and linear probe a ViT-B/16 model on ImageNet via MoCo v3. In this case, SASSL employed a blending factor $\alpha$ uniformly sampled between $0.1$ and $0.5$.

Table \ref{tab:downstream_performance_arch} reports the downstream classification accuracy for ResNet-50 (x4) and ViT-B/16. ResNet-50 (x4) results show SASSL improves top-1 classification accuracy by $1.1\%$, mirroring its earlier improvement. Similarly, ViT-B/16 results show SASSL improves top-1 accuracy by $0.5\%$. These suggest that SASSL is not limited to CNN backbones, but can also be extended to ViTs. While this margin is currently smaller for ViTs, we emphasize that no hyperparameter tuning was employed in these experiments.

\subsection{Ablation Studies}
\label{sec:ablation_study}

To shed light on how SASSL affects accuracy on ImageNet, we break down its components and assess individual contributions to downstream performance. We also study how aligning different layers in the stylization network $\mathcal{T}$ boosts accuracy. See \cref{sec:supp_additional_ablation,sec:supp_computational_requirements} for additional ablations and SASSL's computational requirements.

{\noindent\bf SASSL components.} For the ablation study, we cover four cases: (i) MoCo v2 with default augmentation, (ii) SASSL + MoCo v2 using in-batch representation blending and no pixel interpolation ($\beta=1$), (iii) SASSL + MoCo v2 using in-batch representation blending and pixel interpolation, and (iv) SASSL + MoCo v2 using all its attributes (blending, interpolation and an external style dataset).

Table \ref{tab:ablation} shows our ablation study using MoCo v2 as SSL technique. Results highlight the importance of controlling the amount of stylization using both representation blending and image interpolation. Without image interpolation, using Style Transfer as data augmentation degrades the downstream classification performance by more than $1.5\%$ top-1 accuracy.

On the other hand, by balancing the amount of stylization via blending and interpolation, SASSL boosts performance by more than $1.5\%$. This is a significant improvement for the challenging ImageNet scenario. Finally, by incorporating an external style dataset such as PBN, we further improve downstream task performance by almost $2.1\%$ top-1 accuracy. This shows the importance of diverse style references and their effect on downstream tasks.

\begin{minipage}[t]{0.48\textwidth}
\small
\setlength{\tabcolsep}{2.5pt}
\centering
\captionof{table}{\textbf{Ablation study.} Linear probing accuracy ($\%$) for representations learned via SASSL under different configurations. Mean accuracy reported over five random trials.}
\vspace{-0.15 cm}
\resizebox{\linewidth}{!}{%
\begin{tabular}{ccccc}
\toprule
Method & Configuration & Style & Top-1 Acc. & Top-5 Acc.\\ 
\midrule
  \makecell{MoCo v2\\(Default)}  & $-$ & $-$ & $72.55$ & $91.19$\\
\midrule
  \multirow{8}{*}{\makecell{SASSL +\\MoCo v2 \textbf{(Ours)}}} & \makecell{$p=0.8,$\\ $\alpha \in [0.1, 0.3]$\\ $\beta=1$} & \makecell{ImageNet\\ (\emph{in-batch})} & $70.87$ & $89.33$\\
  \cmidrule{2-5}
  & \makecell{$p=0.8,$\\ $\alpha \in [0.1, 0.3]$\\ $\beta \in [0.1, 0.3]$} & \makecell{ImageNet\\ (\emph{in-batch})} & $74.07$ & $91.58$\\
  \cmidrule{2-5} & \cellcolor{bg_blue}\makecell{$p=0.8,$\\ $\alpha \in [0.1, 0.3]$\\ $\beta \in [0.1, 0.3]$} & \cellcolor{bg_blue}\Gape[0pt][0pt]{\makecell{PBN\\(\emph{external})}} & \cellcolor{bg_blue}$\bf 74.64$ & \cellcolor{bg_blue}$\bf 91.68$\\
\bottomrule
\end{tabular}}
\label{tab:ablation}
\vspace{0.25 cm}
\end{minipage}
\hspace{0.02\textwidth}
\begin{minipage}[t]{0.48\textwidth}
\centering
\setlength{\tabcolsep}{3.5pt}
\small
\captionof{table}{\textbf{Effect of the number of stylized layers in downstream performance.} Linear probing classification accuracy ($\%$) of a ResNet-50 model pretrained via SASSL $+$ MoCo v2, where Style Transfer is applied using a subset of the available layers. Accuracy reported on a single trial.}
\resizebox{\linewidth}{!}{
\begin{tabular}{cccc}
\toprule
Method & Stylized Layers & Top-1 Acc. & Top-5 Acc.\\ 
\midrule
  MoCo v2 (Default) & $-$ & $72.97$ & $90.86$\\
\midrule
  \multirow{6}{*}{\makecell{SASSL +\\ MoCo v2 \textbf{(Ours)}}}
  & None ($\hat{\vz}=\vz_{c}$) & $73.77$ & $91.64$\\
  \cmidrule{2-4}
  & \makecell{First $4$ layers} &$73.75$ & $91.58$\\
  \cmidrule{2-4}
  & \makecell{First $8$ layers} &$74.09$ & $91.76$\\
  \cmidrule{2-4}
  & \makecell{First $10$ layers} & $74.27$ & $91.74$\\
  \cmidrule{2-4}
  & \cellcolor{bg_blue} All ($13$ layers) & \cellcolor{bg_blue} $\bf 75.38$ & \cellcolor{bg_blue}$\bf 92.21$\\
\bottomrule
\end{tabular}}
\label{tab:style_layers_ablation}
\vspace{0.25 cm}
\end{minipage}

{\noindent\bf Number of stylized layers.} 
We explore how the number of layers used to apply style transfer via CIN affects downstream performance. We analyze three cases: (i) stylizing using the first two residual blocks of the Stylization Network $\mathcal{T}$ (four layers from blocks 1 and 2), (ii) the first four residual blocks (eight layers from blocks 1 to 4), and (iii) all five residual blocks (ten layers).

For each case, we pretrain and linearly probe a ResNet-50 on ImageNet using SASSL + MoCo v2 with its recommended settings ($\alpha, \beta \in [0.1, 0.3]$, $p=0.8$) and PBN as style dataset. To fully remove the effect of a style embedding $\vz_{s}$, our comparison includes a model pretrained using the content image itself as style reference ($\hat{\vz}=\vz_{c}$). We also compare our full SASSL + MoCo v2 model, stylizing all residual and upsampling blocks of $\mathcal{T}$ (thirteen layers).

Table \ref{tab:style_layers_ablation} shows a progressive enhancement in accuracy with increasing stylization depth. Adding stylization to the first four layers showed negligible gains, mirroring the accuracy of the unaligned model. Stylizing the first eight and ten layers yielded modest improvements of $0.34\%$ and $0.52\%$, respectively, implying a growing influence of deeper layers on accuracy. Notably, pretraining with full stylization, encompassing both residual and upsampling layers, attains a $1.61\%$ accuracy boost, suggesting the importance of aligning deeper upsampling layers for downstream performance.

\section{Conclusion}
\label{sec:conclusion}
We propose SASSL, a novel data augmentation approach based on Neural Style Transfer that exclusively transforms the style of training samples, diversifying data augmentation during pretraining while preserving semantic attributes. We empirically show our approach outperforms well-established methods such as MoCo v2, SimCLR and BYOL by up to $2\%$ top-1 classification accuracy on ImageNet. SASSL also improves the transfer capabilities of learned representations, enhancing linear probing and fine-tuning performance across domains by up to $10\%$ and $6\%$ top-1 accuracy, respectively. Our technique can be extended to other SSL methods and models with minimum hyperparameter changes, as experimentally shown.

\section*{Broader Impact Statement}
\label{sec:impact_statement}
This work proposes a novel data augmentation approach leveraging Neural Style Transfer to enhance Self-supervised Learning, particularly for domains with limited data or expensive annotations. Our method utilizes semantic-aware image preprocessing to extract robust representations that generalize across diverse domains. This advancement tackles the critical challenge of using unlabeled data for Deep Learning, which has many potential positive impacts in both technical and societal fronts. SASSL's style transfer component helps to reduce sensitivity to image texture, potentially improving model robustness to texture bias. However, since our method primarily modifies data augmentation, it may not fully address other potential biases arising from pre-training datasets or learning strategies.

\section*{Acknowledgements}
The authors would like to thank Arash Afkanpour and Luyang Liu for their insightful comments and feedback.

\bibliography{refs}

\begin{thebibliography}{64}
\providecommand{\natexlab}[1]{#1}
\providecommand{\url}[1]{\texttt{#1}}
\expandafter\ifx\csname urlstyle\endcsname\relax
  \providecommand{\doi}[1]{doi: #1}\else
  \providecommand{\doi}{doi: \begingroup \urlstyle{rm}\Url}\fi

\bibitem[Abadi et~al.(2016)Abadi, Barham, Chen, Chen, Davis, Dean, Devin,
  Ghemawat, Irving, Isard, Kudlur, Levenberg, Monga, Moore, Murray, Steiner,
  Tucker, Vasudevan, Warden, Wicke, Yu, and Zheng]{45381}
Martin Abadi, Paul Barham, Jianmin Chen, Zhifeng Chen, Andy Davis, Jeffrey
  Dean, Matthieu Devin, Sanjay Ghemawat, Geoffrey Irving, Michael Isard,
  Manjunath Kudlur, Josh Levenberg, Rajat Monga, Sherry Moore, Derek~G. Murray,
  Benoit Steiner, Paul Tucker, Vijay Vasudevan, Pete Warden, Martin Wicke, Yuan
  Yu, and Xiaoqiang Zheng.
\newblock Tensorflow: A system for large-scale machine learning.
\newblock In \emph{12th USENIX Symposium on Operating Systems Design and
  Implementation (OSDI 16)}, pp.\  265--283, 2016.
\newblock URL
  \url{https://www.usenix.org/system/files/conference/osdi16/osdi16-abadi.pdf}.

\bibitem[Assran et~al.(2022)Assran, Caron, Misra, Bojanowski, Bordes, Vincent,
  Joulin, Rabbat, and Ballas]{assran_2022_masked}
Mahmoud Assran, Mathilde Caron, Ishan Misra, Piotr Bojanowski, Florian Bordes,
  Pascal Vincent, Armand Joulin, Mike Rabbat, and Nicolas Ballas.
\newblock Masked siamese networks for label-efficient learning.
\newblock In \emph{European Conference on Computer Vision}, pp.\  456--473.
  Springer, 2022.

\bibitem[Bai et~al.(2022)Bai, Yang, Wang, Du, Han, Deng, Wang, and
  Liu]{bai_2022_rsa}
Yingbin Bai, Erkun Yang, Zhaoqing Wang, Yuxuan Du, Bo~Han, Cheng Deng, Dadong
  Wang, and Tongliang Liu.
\newblock Rsa: Reducing semantic shift from aggressive augmentations for
  self-supervised learning.
\newblock \emph{Advances in Neural Information Processing Systems},
  35:\penalty0 21128--21141, 2022.

\bibitem[Bardes et~al.(2021)Bardes, Ponce, and LeCun]{bardes_2021_vicreg}
Adrien Bardes, Jean Ponce, and Yann LeCun.
\newblock Vicreg: Variance-invariance-covariance regularization for
  self-supervised learning.
\newblock \emph{arXiv preprint arXiv:2105.04906}, 2021.

\bibitem[Bossard et~al.(2014)Bossard, Guillaumin, and Van~Gool]{bossard14}
Lukas Bossard, Matthieu Guillaumin, and Luc Van~Gool.
\newblock Food-101 -- mining discriminative components with random forests.
\newblock In \emph{European Conference on Computer Vision}, 2014.

\bibitem[Caron et~al.(2020)Caron, Misra, Mairal, Goyal, Bojanowski, and
  Joulin]{caron_2020_unsupervised}
Mathilde Caron, Ishan Misra, Julien Mairal, Priya Goyal, Piotr Bojanowski, and
  Armand Joulin.
\newblock Unsupervised learning of visual features by contrasting cluster
  assignments.
\newblock \emph{Advances in neural information processing systems},
  33:\penalty0 9912--9924, 2020.

\bibitem[Caron et~al.(2021)Caron, Touvron, Misra, J{\'e}gou, Mairal,
  Bojanowski, and Joulin]{caron_2021_emerging}
Mathilde Caron, Hugo Touvron, Ishan Misra, Herv{\'e} J{\'e}gou, Julien Mairal,
  Piotr Bojanowski, and Armand Joulin.
\newblock Emerging properties in self-supervised vision transformers.
\newblock In \emph{Proceedings of the IEEE/CVF international conference on
  computer vision}, pp.\  9650--9660, 2021.

\bibitem[Chen et~al.(2021{\natexlab{a}})Chen, Wang, Zhang, Zuo, Li, Xing, Lu,
  et~al.]{chen_2021_artistic}
Haibo Chen, Zhizhong Wang, Huiming Zhang, Zhiwen Zuo, Ailin Li, Wei Xing,
  Dongming Lu, et~al.
\newblock Artistic style transfer with internal-external learning and
  contrastive learning.
\newblock \emph{Advances in Neural Information Processing Systems},
  34:\penalty0 26561--26573, 2021{\natexlab{a}}.

\bibitem[Chen et~al.(2020{\natexlab{a}})Chen, Kornblith, Norouzi, and
  Hinton]{chen_2020_simple}
Ting Chen, Simon Kornblith, Mohammad Norouzi, and Geoffrey Hinton.
\newblock A simple framework for contrastive learning of visual
  representations.
\newblock In \emph{International conference on machine learning}, pp.\
  1597--1607. PMLR, 2020{\natexlab{a}}.

\bibitem[Chen et~al.(2021{\natexlab{b}})Chen, Xie, and He]{chen_2021_empirical}
X.~Chen, S.~Xie, and K.~He.
\newblock An empirical study of training self-supervised vision transformers.
\newblock In \emph{2021 IEEE/CVF International Conference on Computer Vision
  (ICCV)}, pp.\  9620--9629, Los Alamitos, CA, USA, oct 2021{\natexlab{b}}.
  IEEE Computer Society.
\newblock \doi{10.1109/ICCV48922.2021.00950}.
\newblock URL
  \url{https://doi.ieeecomputersociety.org/10.1109/ICCV48922.2021.00950}.

\bibitem[Chen \& He(2021)Chen and He]{chen_2021_exploring}
Xinlei Chen and Kaiming He.
\newblock Exploring simple siamese representation learning.
\newblock In \emph{Proceedings of the IEEE/CVF conference on computer vision
  and pattern recognition}, pp.\  15750--15758, 2021.

\bibitem[Chen et~al.(2020{\natexlab{b}})Chen, Fan, Girshick, and
  He]{chen_2020_improved}
Xinlei Chen, Haoqi Fan, Ross Girshick, and Kaiming He.
\newblock Improved baselines with momentum contrastive learning.
\newblock \emph{arXiv preprint arXiv:2003.04297}, 2020{\natexlab{b}}.

\bibitem[Cimpoi et~al.(2014)Cimpoi, Maji, Kokkinos, Mohamed, and
  Vedaldi]{cimpoi_2014_describing}
M.~Cimpoi, S.~Maji, I.~Kokkinos, S.~Mohamed, and A.~Vedaldi.
\newblock Describing textures in the wild.
\newblock In \emph{Proceedings of the {IEEE} Conf. on Computer Vision and
  Pattern Recognition ({CVPR})}, 2014.

\bibitem[Deng et~al.(2009)Deng, Dong, Socher, Li, Li, and
  Fei-Fei]{deng_2009_imagenet}
Jia Deng, Wei Dong, Richard Socher, Li-Jia Li, Kai Li, and Li~Fei-Fei.
\newblock Imagenet: A large-scale hierarchical image database.
\newblock In \emph{2009 IEEE conference on computer vision and pattern
  recognition}, pp.\  248--255. Ieee, 2009.

\bibitem[Devlin et~al.(2018)Devlin, Chang, Lee, and
  Toutanova]{devlin_2018_bert}
Jacob Devlin, Ming-Wei Chang, Kenton Lee, and Kristina Toutanova.
\newblock Bert: Pre-training of deep bidirectional transformers for language
  understanding.
\newblock \emph{arXiv preprint arXiv:1810.04805}, 2018.

\bibitem[Dumoulin et~al.(2017)Dumoulin, Shlens, and
  Kudlur]{dumoulin_2017_learned}
Vincent Dumoulin, Jonathon Shlens, and Manjunath Kudlur.
\newblock A learned representation for artistic style.
\newblock In \emph{International Conference on Learning Representations}, 2017.
\newblock URL \url{https://openreview.net/forum?id=BJO-BuT1g}.

\bibitem[Fei-Fei et~al.(2004)Fei-Fei, Fergus, and Perona]{FeiFei2004LearningGV}
Li~Fei-Fei, Rob Fergus, and Pietro Perona.
\newblock Learning generative visual models from few training examples: An
  incremental bayesian approach tested on 101 object categories.
\newblock \emph{Computer Vision and Pattern Recognition Workshop}, 2004.

\bibitem[Gatys et~al.(2015)Gatys, Ecker, and Bethge]{gatys_2015_neural}
Leon~A Gatys, Alexander~S Ecker, and Matthias Bethge.
\newblock A neural algorithm of artistic style.
\newblock \emph{arXiv preprint arXiv:1508.06576}, 2015.

\bibitem[Gatys et~al.(2016)Gatys, Ecker, and Bethge]{gatys_2016_image}
Leon~A Gatys, Alexander~S Ecker, and Matthias Bethge.
\newblock Image style transfer using convolutional neural networks.
\newblock In \emph{Proceedings of the IEEE Conference on Computer Vision and
  Pattern Recognition}, pp.\  2414--2423, 2016.

\bibitem[Gatys et~al.(2017)Gatys, Ecker, Bethge, Hertzmann, and
  Shechtman]{gatys_2017_controlling}
Leon~A Gatys, Alexander~S Ecker, Matthias Bethge, Aaron Hertzmann, and Eli
  Shechtman.
\newblock Controlling perceptual factors in neural style transfer.
\newblock In \emph{Proceedings of the IEEE Conference on Computer Vision and
  Pattern Recognition}, pp.\  3985--3993, 2017.

\bibitem[Geirhos et~al.(2018)Geirhos, Rubisch, Michaelis, Bethge, Wichmann, and
  Brendel]{geirhos_2018_imagenet}
Robert Geirhos, Patricia Rubisch, Claudio Michaelis, Matthias Bethge, Felix~A
  Wichmann, and Wieland Brendel.
\newblock Imagenet-trained cnns are biased towards texture; increasing shape
  bias improves accuracy and robustness.
\newblock \emph{arXiv preprint arXiv:1811.12231}, 2018.

\bibitem[Grill et~al.(2020)Grill, Strub, Altch{\'e}, Tallec, Richemond,
  Buchatskaya, Doersch, Avila~Pires, Guo, Gheshlaghi~Azar,
  et~al.]{grill_2020_bootstrap}
Jean-Bastien Grill, Florian Strub, Florent Altch{\'e}, Corentin Tallec, Pierre
  Richemond, Elena Buchatskaya, Carl Doersch, Bernardo Avila~Pires, Zhaohan
  Guo, Mohammad Gheshlaghi~Azar, et~al.
\newblock Bootstrap your own latent-a new approach to self-supervised learning.
\newblock \emph{Advances in neural information processing systems},
  33:\penalty0 21271--21284, 2020.

\bibitem[Han et~al.(2022)Han, Fang, Li, Hong, Armin, Reid, Petersson, and
  Li]{han_2022_you}
Junlin Han, Pengfei Fang, Weihao Li, Jie Hong, Mohammad~Ali Armin, Ian Reid,
  Lars Petersson, and Hongdong Li.
\newblock You only cut once: Boosting data augmentation with a single cut.
\newblock In \emph{International Conference on Machine Learning}, pp.\
  8196--8212. PMLR, 2022.

\bibitem[He et~al.(2016)He, Zhang, Ren, and Sun]{He_2016_CVPR}
Kaiming He, Xiangyu Zhang, Shaoqing Ren, and Jian Sun.
\newblock Deep residual learning for image recognition.
\newblock In \emph{Proceedings of the IEEE Conference on Computer Vision and
  Pattern Recognition (CVPR)}, June 2016.

\bibitem[He et~al.(2020)He, Fan, Wu, Xie, and Girshick]{he_2020_momentum}
Kaiming He, Haoqi Fan, Yuxin Wu, Saining Xie, and Ross Girshick.
\newblock Momentum contrast for unsupervised visual representation learning.
\newblock In \emph{Proceedings of the IEEE/CVF conference on computer vision
  and pattern recognition}, pp.\  9729--9738, 2020.

\bibitem[He et~al.(2022)He, Chen, Xie, Li, Doll{\'a}r, and
  Girshick]{he_2022_masked}
Kaiming He, Xinlei Chen, Saining Xie, Yanghao Li, Piotr Doll{\'a}r, and Ross
  Girshick.
\newblock Masked autoencoders are scalable vision learners.
\newblock In \emph{Proceedings of the IEEE/CVF conference on computer vision
  and pattern recognition}, pp.\  16000--16009, 2022.

\bibitem[Heeger \& Bergen(1995)Heeger and Bergen]{heeger_1995_pyramid}
David~J Heeger and James~R Bergen.
\newblock Pyramid-based texture analysis/synthesis.
\newblock In \emph{Proceedings of the Conference on Computer Graphics and
  Interactive Techniques}, pp.\  229--238, 1995.

\bibitem[Heitz et~al.(2021)Heitz, Vanhoey, Chambon, and
  Belcour]{heitz_2021_sliced}
Eric Heitz, Kenneth Vanhoey, Thomas Chambon, and Laurent Belcour.
\newblock A sliced {W}asserstein loss for neural texture synthesis.
\newblock In \emph{Proceedings of the IEEE/CVF Conference on Computer Vision
  and Pattern Recognition}, pp.\  9412--9420, 2021.

\bibitem[Hong et~al.(2021)Hong, Choi, and Kim]{hong_2021_stylemix}
Minui Hong, Jinwoo Choi, and Gunhee Kim.
\newblock Stylemix: Separating content and style for enhanced data
  augmentation.
\newblock In \emph{Proceedings of the IEEE/CVF conference on computer vision
  and pattern recognition}, pp.\  14862--14870, 2021.

\bibitem[Huang \& Belongie(2017)Huang and Belongie]{huang_2017_arbitrary}
Xun Huang and Serge Belongie.
\newblock Arbitrary style transfer in real-time with adaptive instance
  normalization.
\newblock In \emph{Proceedings of the IEEE International Conference on Computer
  Vision}, pp.\  1501--1510, 2017.

\bibitem[{iNaturalist} 2021()]{inaturalist21}
{iNaturalist} 2021.
\newblock {iNaturalist} 2021 competition dataset.
\newblock ~\url{https://github.com/visipedia/inat_comp/tree/master/2021}, 2021.

\bibitem[Jackson et~al.(2019)Jackson, Abarghouei, Bonner, Breckon, and
  Obara]{jackson_2019_style}
Philip~TG Jackson, Amir~Atapour Abarghouei, Stephen Bonner, Toby~P Breckon, and
  Boguslaw Obara.
\newblock Style augmentation: data augmentation via style randomization.
\newblock In \emph{CVPR workshops}, volume~6, pp.\  10--11, 2019.

\bibitem[Jing et~al.(2019)Jing, Yang, Feng, Ye, Yu, and Song]{jing_2019_neural}
Yongcheng Jing, Yezhou Yang, Zunlei Feng, Jingwen Ye, Yizhou Yu, and Mingli
  Song.
\newblock Neural style transfer: A review.
\newblock \emph{IEEE transactions on visualization and computer graphics},
  26\penalty0 (11):\penalty0 3365--3385, 2019.

\bibitem[Jing et~al.(2020)Jing, Liu, Ding, Wang, Ding, Song, and
  Wen]{jing_2020_dynamic}
Yongcheng Jing, Xiao Liu, Yukang Ding, Xinchao Wang, Errui Ding, Mingli Song,
  and Shilei Wen.
\newblock Dynamic instance normalization for arbitrary style transfer.
\newblock In \emph{Proceedings of the AAAI Conference on Artificial
  Intelligence}, volume~34, pp.\  4369--4376, 2020.

\bibitem[Johnson et~al.(2016)Johnson, Alahi, and
  Fei-Fei]{johnson_2016_perceptual}
Justin Johnson, Alexandre Alahi, and Li~Fei-Fei.
\newblock Perceptual losses for real-time style transfer and super-resolution.
\newblock In \emph{European conference on computer vision}, pp.\  694--711.
  Springer, 2016.

\bibitem[Kaggle \& EyePacs(2015)Kaggle and
  EyePacs]{kaggle_diabetic_retinopathy}
Kaggle and EyePacs.
\newblock Kaggle diabetic retinopathy detection, jul 2015.
\newblock URL
  \url{https://www.kaggle.com/c/diabetic-retinopathy-detection/data}.

\bibitem[Kan(2016)]{painter_by_numbers}
Wendy Kan.
\newblock Painter by numbers, 2016.
\newblock URL \url{https://kaggle.com/competitions/painter-by-numbers}.

\bibitem[Kessy et~al.(2018)Kessy, Lewin, and Strimmer]{kessy_2018_optimal}
Agnan Kessy, Alex Lewin, and Korbinian Strimmer.
\newblock Optimal whitening and decorrelation.
\newblock \emph{The American Statistician}, 72\penalty0 (4):\penalty0 309--314,
  2018.

\bibitem[Kornblith et~al.(2019)Kornblith, Shlens, and
  Le]{kornblith_2019_better}
Simon Kornblith, Jonathon Shlens, and Quoc~V Le.
\newblock Do better imagenet models transfer better?
\newblock In \emph{Proceedings of the IEEE/CVF conference on computer vision
  and pattern recognition}, pp.\  2661--2671, 2019.

\bibitem[Krause et~al.(2013)Krause, Stark, Deng, and
  Fei-Fei]{KrauseStarkDengFei-Fei_3DRR2013}
Jonathan Krause, Michael Stark, Jia Deng, and Li~Fei-Fei.
\newblock 3d object representations for fine-grained categorization.
\newblock In \emph{4th International IEEE Workshop on 3D Representation and
  Recognition (3dRR-13)}, Sydney, Australia, 2013.

\bibitem[Krizhevsky(2009)]{Krizhevsky09learningmultiple}
Alex Krizhevsky.
\newblock Learning multiple layers of features from tiny images.
\newblock Technical report, 2009.

\bibitem[Lee et~al.(2021)Lee, Lee, Lee, Lee, and Shin]{lee_2021_improving}
Hankook Lee, Kibok Lee, Kimin Lee, Honglak Lee, and Jinwoo Shin.
\newblock Improving transferability of representations via augmentation-aware
  self-supervision.
\newblock \emph{Advances in Neural Information Processing Systems},
  34:\penalty0 17710--17722, 2021.

\bibitem[Li et~al.(2017{\natexlab{a}})Li, Wang, Liu, and
  Hou]{li_2017_demystifying}
Yanghao Li, Naiyan Wang, Jiaying Liu, and Xiaodi Hou.
\newblock Demystifying neural style transfer.
\newblock \emph{arXiv preprint arXiv:1701.01036}, 2017{\natexlab{a}}.

\bibitem[Li et~al.(2017{\natexlab{b}})Li, Fang, Yang, Wang, Lu, and
  Yang]{li_2017_universal}
Yijun Li, Chen Fang, Jimei Yang, Zhaowen Wang, Xin Lu, and Ming-Hsuan Yang.
\newblock Universal style transfer via feature transforms.
\newblock In \emph{Proceedings of the 31st International Conference on Neural
  Information Processing Systems}, NIPS'17, pp.\  385--395, Red Hook, NY, USA,
  2017{\natexlab{b}}. Curran Associates Inc.
\newblock ISBN 9781510860964.

\bibitem[Liu et~al.(2021)Liu, Lin, He, Li, Wang, Li, Sun, Li, and
  Ding]{liu_2021_adaattn}
Songhua Liu, Tianwei Lin, Dongliang He, Fu~Li, Meiling Wang, Xin Li, Zhengxing
  Sun, Qian Li, and Errui Ding.
\newblock Adaattn: Revisit attention mechanism in arbitrary neural style
  transfer.
\newblock In \emph{Proceedings of the IEEE/CVF International Conference on
  Computer Vision}, pp.\  6649--6658, 2021.

\bibitem[Nilsback \& Zisserman(2008)Nilsback and Zisserman]{Nilsback08}
M-E. Nilsback and A.~Zisserman.
\newblock Automated flower classification over a large number of classes.
\newblock In \emph{Proceedings of the Indian Conference on Computer Vision,
  Graphics and Image Processing}, Dec 2008.

\bibitem[Portilla \& Simoncelli(2000)Portilla and
  Simoncelli]{portilla_2000_parametric}
Javier Portilla and Eero~P Simoncelli.
\newblock A parametric texture model based on joint statistics of complex
  wavelet coefficients.
\newblock \emph{International Journal of Computer Vision}, 40\penalty0
  (1):\penalty0 49--70, 2000.

\bibitem[Purushwalkam \& Gupta(2020)Purushwalkam and
  Gupta]{purushwalkam_2020_demystifying}
Senthil Purushwalkam and Abhinav Gupta.
\newblock Demystifying contrastive self-supervised learning: Invariances,
  augmentations and dataset biases.
\newblock \emph{Advances in Neural Information Processing Systems},
  33:\penalty0 3407--3418, 2020.

\bibitem[Reed et~al.(2021)Reed, Metzger, Srinivas, Darrell, and
  Keutzer]{reed_2021_selfaugment}
Colorado~J Reed, Sean Metzger, Aravind Srinivas, Trevor Darrell, and Kurt
  Keutzer.
\newblock Selfaugment: Automatic augmentation policies for self-supervised
  learning.
\newblock In \emph{Proceedings of the IEEE/CVF conference on computer vision
  and pattern recognition}, pp.\  2674--2683, 2021.

\bibitem[Risser et~al.(2017)Risser, Wilmot, and Barnes]{risser_2017_stable}
Eric Risser, Pierre Wilmot, and Connelly Barnes.
\newblock Stable and controllable neural texture synthesis and style transfer
  using histogram losses.
\newblock 2017.
\newblock URL \url{http://arxiv.org/abs/1701.08893}.

\bibitem[Sheng et~al.(2018)Sheng, Lin, Shao, and Wang]{sheng_2018_avatar}
Lu~Sheng, Ziyi Lin, Jing Shao, and Xiaogang Wang.
\newblock Avatar-net: Multi-scale zero-shot style transfer by feature
  decoration.
\newblock In \emph{Proceedings of the IEEE Conference on Computer Vision and
  Pattern Recognition}, pp.\  8242--8250, 2018.

\bibitem[Simonyan \& Zisserman(2014)Simonyan and Zisserman]{simonyan_2014_very}
Karen Simonyan and Andrew Zisserman.
\newblock Very deep convolutional networks for large-scale image recognition.
\newblock \emph{arXiv preprint arXiv:1409.1556}, 2014.

\bibitem[Szegedy et~al.(2016)Szegedy, Vanhoucke, Ioffe, Shlens, and
  Wojna]{szegedy_2016_rethinking}
Christian Szegedy, Vincent Vanhoucke, Sergey Ioffe, Jon Shlens, and Zbigniew
  Wojna.
\newblock Rethinking the inception architecture for computer vision.
\newblock In \emph{Proceedings of the IEEE conference on computer vision and
  pattern recognition}, pp.\  2818--2826, 2016.

\bibitem[Tian et~al.(2020)Tian, Sun, Poole, Krishnan, Schmid, and
  Isola]{tian_2020_makes}
Yonglong Tian, Chen Sun, Ben Poole, Dilip Krishnan, Cordelia Schmid, and
  Phillip Isola.
\newblock What makes for good views for contrastive learning?
\newblock \emph{Advances in neural information processing systems},
  33:\penalty0 6827--6839, 2020.

\bibitem[Van~der Maaten \& Hinton(2008)Van~der Maaten and
  Hinton]{van_2008_visualizing}
Laurens Van~der Maaten and Geoffrey Hinton.
\newblock Visualizing data using t-sne.
\newblock \emph{Journal of machine learning research}, 9\penalty0 (11), 2008.

\bibitem[Wagner et~al.(2022)Wagner, Ferreira, Stoll, Schirrmeister, M{\"u}ller,
  and Hutter]{wagner_2022_importance}
Diane Wagner, Fabio Ferreira, Danny Stoll, Robin~Tibor Schirrmeister, Samuel
  M{\"u}ller, and Frank Hutter.
\newblock On the importance of hyperparameters and data augmentation for
  self-supervised learning.
\newblock \emph{arXiv preprint arXiv:2207.07875}, 2022.

\bibitem[Wang et~al.(2020)Wang, Zhao, Chen, Qiu, Mo, Lin, Xing, and
  Lu]{wang_2020_diversified}
Zhizhong Wang, Lei Zhao, Haibo Chen, Lihong Qiu, Qihang Mo, Sihuan Lin, Wei
  Xing, and Dongming Lu.
\newblock Diversified arbitrary style transfer via deep feature perturbation.
\newblock In \emph{Proceedings of the IEEE/CVF Conference on Computer Vision
  and Pattern Recognition}, pp.\  7789--7798, 2020.

\bibitem[{Xiao} et~al.(2010){Xiao}, {Hays}, {Ehinger}, {Oliva}, and
  {Torralba}]{Xiao:2010}
J.~{Xiao}, J.~{Hays}, K.~A. {Ehinger}, A.~{Oliva}, and A.~{Torralba}.
\newblock Sun database: Large-scale scene recognition from abbey to zoo.
\newblock In \emph{2010 IEEE Computer Society Conference on Computer Vision and
  Pattern Recognition}, pp.\  3485--3492, June 2010.
\newblock \doi{10.1109/CVPR.2010.5539970}.

\bibitem[Yoo et~al.(2019)Yoo, Uh, Chun, Kang, and Ha]{yoo_2019_photorealistic}
Jaejun Yoo, Youngjung Uh, Sanghyuk Chun, Byeongkyu Kang, and Jung-Woo Ha.
\newblock Photorealistic style transfer via wavelet transforms.
\newblock In \emph{Proceedings of the IEEE International Conference on Computer
  Vision}, pp.\  9036--9045, 2019.

\bibitem[You et~al.(2017)You, Gitman, and Ginsburg]{you2017large}
Yang You, Igor Gitman, and Boris Ginsburg.
\newblock Large batch training of convolutional networks.
\newblock \emph{arXiv preprint arXiv:1708.03888}, 2017.

\bibitem[Zbontar et~al.(2021)Zbontar, Jing, Misra, LeCun, and
  Deny]{zbontar2021barlow}
Jure Zbontar, Li~Jing, Ishan Misra, Yann LeCun, and St{\'e}phane Deny.
\newblock Barlow twins: Self-supervised learning via redundancy reduction.
\newblock In \emph{International Conference on Machine Learning}, pp.\
  12310--12320. PMLR, 2021.

\bibitem[Zhang et~al.(2018)Zhang, Isola, Efros, Shechtman, and
  Wang]{zhang_2018_unreasonable}
Richard Zhang, Phillip Isola, Alexei~A Efros, Eli Shechtman, and Oliver Wang.
\newblock The unreasonable effectiveness of deep features as a perceptual
  metric.
\newblock In \emph{Proceedings of the IEEE Conference on Computer Vision and
  Pattern Recognition}, pp.\  586--595, 2018.

\bibitem[Zheng et~al.(2019)Zheng, Chalasani, Ghosal, Lutz, and
  Smolic]{zheng_2019_stada}
Xu~Zheng, Tejo Chalasani, Koustav Ghosal, Sebastian Lutz, and Aljosa Smolic.
\newblock Stada: Style transfer as data augmentation.
\newblock \emph{arXiv preprint arXiv:1909.01056}, 2019.

\bibitem[Zhu et~al.(2000)Zhu, Liu, and Wu]{zhu_2000_exploring}
Song~Chun Zhu, Xiu~Wen Liu, and Ying~Nian Wu.
\newblock Exploring texture ensembles by efficient {M}arkov chain monte
  carlo-toward a ``trichromacy'' theory of texture.
\newblock \emph{IEEE Transactions on Pattern Analysis and Machine
  Intelligence}, 22\penalty0 (6):\penalty0 554--569, 2000.

\end{thebibliography}
\bibliographystyle{tmlr}

\newpage
\appendix
\onecolumn
\section{Appendix}
The Appendix is organized as follows:
\begin{itemize}[leftmargin=0.45cm,topsep=0pt]
    \item Section~\ref{sec:supp_datasets} shows the details of the target and style datasets used in our downstream evaluations.
    \item Section~\ref{sec:supp_style_visualization} provides insight on how to select an external style dataset by analyzing their style representations.
    \item Section~\ref{sec:supp_additional_experimental_details} includes additional information on the pretraining settings used in our experiments.
    \item Section~\ref{sec:supp_downstream_settings} includes detailed information on the downstream task settings used in our experiments.
    \item Section~\ref{sec:supp_additional_mocov2_results} reports full downstream classification accuracy (mean and standard deviation) on our SASSL + MoCo v2 experiments.
    \item Section~\ref{sec:supp_additional_ablation} includes an additional ablation study to better understand the effect of Neural Style Transfer in the downstream performance.
    \item Section~\ref{sec:supp_computational_requirements} covers the computational requirements of our proposed method.
\end{itemize}

\subsection{Target and Style Datasets}
\label{sec:supp_datasets}
We provide the details of the image datasets used in our experiments. Table \ref{tab:supp_datasets} covers both target and style datasets, including their size, splits and number of classes.

\begin{table}[h]
\centering
\small
\setlength{\tabcolsep}{2.5pt}
\centering
\caption{\textbf{Target and Style Datasets.} Additional details on number of classes, data split and samples of the image datasets used in our experiments.}
\resizebox{\linewidth}{!}{%
\begin{tabular}{cccccc}
\toprule
Dataset & Task & Classes & Train Split & Val. Split & Test Split\\ 
\midrule
ImageNet \cite{deng_2009_imagenet}  & Pretraining, Target, Style & $1,000$ & $1,281,167$ & $-$ & $50,000$\\
iNaturalist `21 \citep{inaturalist21} & Target, Style & $10,000$ & $2,686,843$ & $-$ & $500,000$\\
Diabetic Retinopathy Detection \citep{kaggle_diabetic_retinopathy}& Target, Style & $5$ & $35,126$ & $10,906$ & $42,670$\\
Describable Textures Dataset \citep{cimpoi_2014_describing}& Target, Style & $47$ & $1,880$ & $1,880$ & $1,880$\\
Painter by Numbers \citep{painter_by_numbers}& Style & $1,584$ & $79,433$ & $-$ & $23,817$\\
ImageNet $1\%$ \citep{chen_2020_simple}& Target & $1,000$ & $12,811$ & $-$ & $50,000$\\
Food101 \citep{bossard14}& Target & $101$ & $75,750$ & $-$ & $25,250$\\
CIFAR10 \citep{Krizhevsky09learningmultiple}& Target & $10$ & $50,000$ & $-$ & $10,000$\\
CIFAR100 \citep{Krizhevsky09learningmultiple}& Target & $100$ & $50,000$ & $-$ & $10,000$\\
SUN397 \citep{Xiao:2010}& Target & $397$ & $76,128$ & $10,875$ & $21,750$\\
Cars \citep{KrauseStarkDengFei-Fei_3DRR2013}& Target & $196$ & $8,144$ & $-$ & $8,041$\\
Caltech-101 \citep{FeiFei2004LearningGV}& Target & $102$ & $3,060$ & $-$ & $6,084$\\
Flowers \citep{Nilsback08}& Target & $102$ & $1,020$ & $1,020$ & $6,149$ \\
\bottomrule
\end{tabular}}
\label{tab:supp_datasets}
\end{table}

\subsection{Style Dataset Selection}
\label{sec:supp_style_visualization}

Our transfer learning results in Section \ref{sec:experiments} demonstrate that SASSL achieves improved or on-par downstream performance across multiple datasets by incorporating Neural Style Transfer (NST) as data augmentation. This raises an important question: how can we select an external style dataset to ensure downstream accuracy improvement? Here, we delve into the similarity between datasets in terms of their styles and establish its connection to the performance improvement gained by using them as external style references.

We focus on the linear-probing scenario as it freezes the representation model, forcing the classification head to rely on the representations learned during pretraining (rather than updating them as is the case with fine-tuning) to distinguish between the target categories.

As an example, in our linear probing experiments presented in Table \ref{tab:transfer_learning_full}, when Diabetic Retinopathy is used as the target dataset, the downstream accuracy achieved via SASSL + MoCo v2 is comparable to that of the default MoCo v2 algorithm, meaning there is no improvement in performance. We hypothesize that this is because the style representations of Diabetic Retinopathy are significantly different from those of the pretraining (ImageNet) and style datasets. Therefore, learning representations that are invariant to such a distinct set of styles does not contribute to distinguishing between target classes.

To support our hypothesis, we visualize the relationship between style representations using low-dimensional embeddings generated via t-SNE \citep{van_2008_visualizing} to capture the similarity between styles of different datasets. Style representations of each dataset, corresponding to vectors of length $100$, are obtained using an InceptionV3 feature extractor, as done by the \textit{Fast Style Transfer} algorithm. Next, we randomly select 1,800 style representations from each dataset and compute their two-dimensional embeddings using a perplexity of $30$, early exaggeration of $12$, and initializing the dimensionality reduction using PCA. We compute embeddings using 2,048 iterations.

\begin{figure*}[t]
    \centering
    \subfloat{\includegraphics[width=\linewidth]{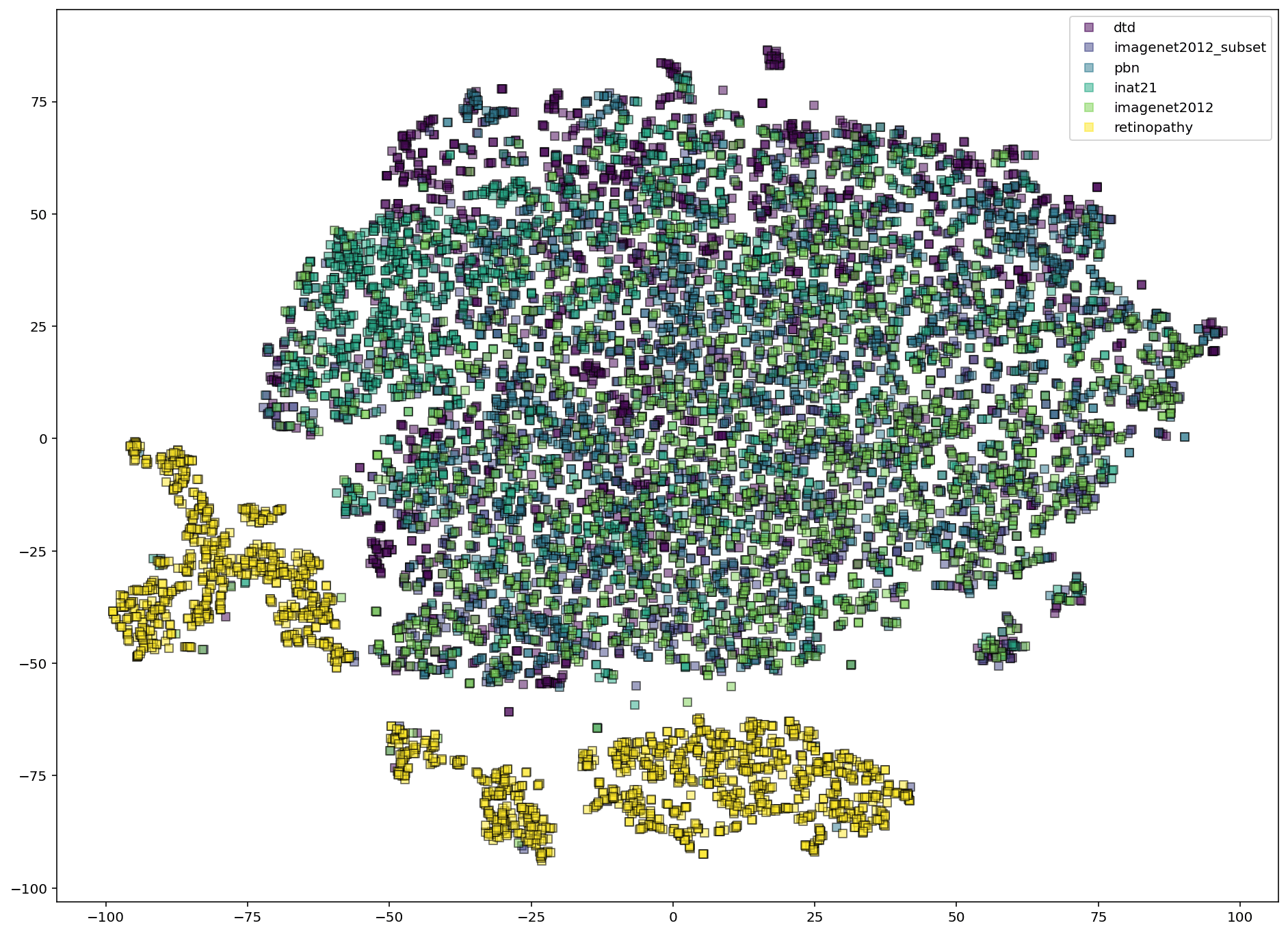}\vspace{-0.5\baselineskip}}
    \caption{\textbf{t-SNE visualization of style representations.} Two-dimensional embeddings of the style representations of different datasets, extracted by the \textit{Fast Style Transfer} method. Style embeddings of the Diabetic Retinopathy dataset (marked in yellow) form clusters that do not overlap with the rest of datasets, while embeddings from the remaining datasets are close to each other.}
    \label{fig:supp_tsne}
    \vspace{-0.3cm}
\end{figure*}

Figure \ref{fig:supp_tsne} depicts the low-dimensional embeddings obtained via t-SNE from six diverse datasets used in our transfer learning experiments. The low-dimensional representation shows that the style embeddings from Diabetic Retinopathy are significantly distinct from those of the rest of datasets, including ImageNet. This aligns with our hypothesis, suggesting that SASSL improves transfer learning when the pretraining and style references are similar to those of the target dataset. From the perspective of t-SNE embeddings, this implies that pretraining and style datasets must have a good overlap with the target dataset for SASSL to improve downstream performance.

\subsection{Additional Experimental Details}
\label{sec:supp_additional_experimental_details}
All of our experiments were implemented in Tensorflow \citep{45381}. All our models were pretrained on $64$ TPUs using a batch size of $4,096$. Both the MoCo v2 models pretrained using the default augmentation and our proposed SASSL approach do not use a dictionary queue.

We used a ResNet-50 \citep{He_2016_CVPR} as our representation backbone. For our projection head, we used a Multilayer perceptron with $4,096$ hidden features, and an output dimensionality of $256$. Our left tower used a prediction network with the same architecture as the projector, similar to the setup used in BYOL \citep{grill_2020_bootstrap}. Our right tower was a momentum encoder, having the same encoder and projector as the left tower, but whose parameters were an exponential moving average of the corresponding parameters in the left tower and were not trained via gradient descent. Similar to previous works, we used a momentum which started at $0.996$, and which followed a cosine decay schedule ending at 1.0. For the pretraining loss, we used the InfoNCE loss with a temperature of 0.1, similar to what was done in both MoCo v2 and SimCLR.

For our pretraining augmentations, we followed the setup used in BYOL \citep{grill_2020_bootstrap}. The operations used and their hyperparameters (in order of application) are as follows:
\begin{enumerate}[leftmargin=0.45cm]
    \item Random cropping and rescaling to $224\times224$ with the area chosen randomly between $0.08$ and $1.0$ of the original image and with a logarithmically distributed axis ratio between $3/4$ and $4/3$. This was applied with a probability of 1.0 since it was necessary to get a fixed image shape.
    \item Random horizontal flipping with a probability of $0.5$ that it will be applied.
    \item Random color jitter. Color jitter consists of $4$ independent transformations, each of which is applied in a random order with randomly chosen values. This transform is described in greater detail in \citet{chen_2020_simple} and \citet{grill_2020_bootstrap}. We used the same configuration as used in those papers.
    \item Random grayscaling with a probability of 0.2 that it will be applied.
    \item Random blurring with a kernel width distributed randomly between 0.1 and 2.0 pixels. Similar to \citet{grill_2020_bootstrap}, we used a probability of $1.0$ for the left view, and a probability of $0.1$ for the right view.
    \item Random solarization, which was only applied to the right view with a threshold of $0.5$ and a probability of $0.2$ that it will be applied.
\end{enumerate}
When using SASSL, we applied NST after random cropping and before random horizontal flipping. Our default hyperparameters for SASSL were blending and interpolation factors drawn randomly from a uniform distribution between $0.1$ and $0.3$, and a probability of $0.8$ that NST will be applied.

For optimization, we used the LARS optimizer \citep{you2017large} with a cosine decayed learning rate warmed up to $4.8$ over the course of the first $10$ epochs. Similar to previous works, we used a trust coefficient of $0.001$, exempted biases and batchnorm parameters from layer adaptation and weight decay, and used a weight decay of $1.5\times10^{-6}$.

\begin{table}
\centering
\small
\rowcolors{2}{bg_blue}{}
\centering
\caption{\textbf{SASSL $+$ MoCo v2 downstream classification accuracy on ImageNet}. Linear probing accuracy of ResNet-50 pretrained via SASSL $+$ MoCo v2. Mean and standard deviation reported over five random trials.}
\vspace{-0.2 cm}
\begin{tabular}{cccccc}
\toprule
Method & Top-1 Acc. $(\%)$ & Top-5 Acc. $(\%)$\\ 
\midrule
    \makecell{MoCo v2 (Default)} & $72.55 \pm 0.67$ & $91.19 \pm 0.34$\\
    \makecell{SASSL $+$ MoCo v2 \textbf{(Ours)}} &   $\bf 74.64 \pm 0.43$ & $\bf 91.68 \pm 0.36$\\
\bottomrule
\end{tabular}
\label{tab:supp_downstream_performance_mocov2}
\vspace{-0.25 cm}
\end{table}
\begin{table*}
\centering
\small
\caption{\textbf{Additional experiments on transfer learning.} Downstream top-1 classification accuracy of ResNet-50 pretrained via MoCo v2 $+$ SASSL on ImageNet. Accuracy evaluated on six out of twelve target datasets. SASSL generates specialized representations that improve transfer learning performance, both in linear probing and fine-tuning. Mean and standard deviation reported over five random trials.}
\resizebox{\linewidth}{!}{%
\begin{tabular}{cccccccc}
\toprule
& & \multicolumn{5}{c}{\textbf{Target Dataset}}\\
& & ImageNet & ImageNet (1\%) & iNat21 & Retinopathy & DTD & Food101\\
\midrule
& & \multicolumn{5}{c}{\emph{Linear Probing}}\\
\cmidrule{2-8}
  & \makecell{None (Default)}  & $72.55 \pm 0.67$ & $53.23 \pm 0.45$ & $41.33 \pm 0.2$ & $\bf75.88 \pm 0.12$ & $72.68 \pm 0.7$ & $ 73.82 \pm 0.1$ \\
  \rowcolor{bg_blue}\cellcolor{white}& \textcolor{black}{ImageNet (\textbf{Ours})} & $74.07 \pm 0.46$ & $56.87 \pm 0.43$ & $45.01 \pm 0.04$ & $75.75 \pm 0.1$ & $73.69 \pm 1.22$ & $74.43 \pm 0.38$ \\
  \rowcolor{bg_blue}\cellcolor{white}& \textcolor{black}{iNat21 (\textbf{Ours})} & $74.28 \pm 0.38$ & $56.76 \pm 0.23$ & $44.70 \pm 0.37$ & $75.75 \pm 0.17$ & $72.75 \pm 1.01$ & $ 74.3\pm 0.48$ \\
  \rowcolor{bg_blue}\cellcolor{white}& \textcolor{black}{Retinopathy (\textbf{Ours})} & $74.02 \pm 0.61$ & $\bf 56.99 \pm 0.26$ & $44.9 \pm 0.16$ & $75.78 \pm 0.08$ & $73.73 \pm 0.57$ & $74.53 \pm 0.29$ \\
  \rowcolor{bg_blue}\cellcolor{white}& \textcolor{black}{DTD (\textbf{Ours})} & $74.32 \pm 0.37$ & $56.77 \pm 0.36$ & $\bf45.08 \pm 0.31$ & $75.76 \pm 0.11$ & $\bf74.41 \pm 1.39$ & $\bm{74.88 \pm 0.32}$ \\
  \rowcolor{bg_blue}\cellcolor{white}& \textcolor{black}{PBN (\textbf{Ours})} & $\bf74.64 \pm 0.43$ & $56.9 \pm 0.18$ & $ 45.02 \pm 0.14$ & $75.79 \pm 0.07$ & $72.77 \pm 0.77$ & $74.37 \pm 0.19$ \\
\cmidrule{2-8}
& & \multicolumn{5}{c}{\emph{Fine-tuning}}\\
\cmidrule{2-8}
& \makecell{None (Default)}  & $74.89 \pm 0.67$ & $51.61 \pm 0.13$ & $77.92 \pm 0.14$ & $78.89 \pm 0.2$ & $71.54 \pm 0.43$ & $87.25 \pm 0.11$\\
  \rowcolor{bg_blue}\cellcolor{white}& \textcolor{black}{ImageNet (\textbf{Ours})} & $75.52 \pm 0.23$ & $51.74 \pm 0.14$ & $79.21 \pm 0.07$ & $79.64 \pm 0.16$ & $\bf 72.31 \pm 1.85$ & $ 87.48\pm 0.21$\\
  \rowcolor{bg_blue}\cellcolor{white}& \textcolor{black}{iNat21 (\textbf{Ours})} & $\bf 75.58 \pm 0.47$ & $\bf 51.86 \pm 0.3$ & $79.19 \pm 0.12$ & $79.6 \pm 0.23$ & $71.35 \pm 1.58$ & $87.4 \pm 0.38$ \\
  \rowcolor{bg_blue}\cellcolor{white}& \textcolor{black}{Retinopathy (\textbf{Ours})} & $75.52 \pm 0.64$ & $51.76 \pm 0.26$ & $79.23 \pm 0.05$ & $79.63 \pm 0.13$ & $72.07 \pm 1.61$ & $87.39 \pm 0.19$ \\
  \rowcolor{bg_blue}\cellcolor{white}& \textcolor{black}{DTD (\textbf{Ours})} & $75.24 \pm 0.65$ & $51.73 \pm 0.19$ & $\bf 79.24 \pm 0.08$ & $\bf 79.7 \pm 0.15$ & $70.59 \pm 1.42$ & $\bm{87.66 \pm 0.23}$ \\
  \rowcolor{bg_blue}\cellcolor{white} \multirow{-14}{*}{\rotatebox[origin=c]{90}{\textbf{Style Dataset}}} & \textcolor{black}{PBN (\textbf{Ours})} & $75.05 \pm 0.69$ & $51.85 \pm 0.16$ & $ 79.2 \pm 0.1$ & $79.63 \pm 0.13$ & $71.35 \pm 0.96$ & $87.56 \pm 0.38$ \\
\bottomrule
\end{tabular}}
\label{tab:supp_transfer_learning01}
\end{table*}

\subsection{Downstream Training and Testing Settings}
\label{sec:supp_downstream_settings}
For performance evaluation on downstream tasks, all our models were trained on $64$ TPUs, but using a batch size of $1,024$. In this section, we provide additional details of the downstream training configuration used in our experiments. These cover data augmentation, optimizer and scheduler settings for both linear probing and fine-tuning scenarios.

{\noindent \bf Linear Probing Settings.} We base our linear probing settings on those used by well-established SSL methods \citep{grill_2020_bootstrap, chen_2020_simple, kornblith_2019_better} with some changes on the optimizer settings. We also adapt the augmentation pipeline based on the target dataset.

In all our linear probing experiments, the optimization method corresponds to SGD with Nesterov momentum using a momentum parameter of $0.9$. We use an initial learning rate of $0.2$ and no weight decay. We use a cosine scheduler with no warmup epochs and a decay factor of $10^{-6}$. Similarly to previous work, for datasets including a validation split, we trained the linear probe on the training and validation splits together, and evaluated on the testing set.

For small target datasets (ImageNet $1\%$, Retinopathy, and DTD), models were trained for $5,000$ iterations using a batch size of $1,024$, which is consistent with the $20,000$ iterations using a batch size of $256$ reported by previous methods. No data augmentation is applied during training. Instead, during both training and testing, images are resized to $224$ pixels along the shorter dimension followed by a $224 \times 224$ center crop and then standardized using the ImageNet statistics. 

For iNat21, comprised by $2.6$ million training images, we train the linear probe for $90$ epochs. We empirically found that longer training significantly improved the downstream classification performance both for our proposed SASSL pipeline as well as the default augmentation pipeline.

Similarly, for ImageNet, comprised by $1.2$ million training images, we also train the linear probe for $90$ epochs. Additionally, we included random cropping, horizontal flipping and color augmentations (grayscale, solarization and blurring) during training.

{\noindent \bf Fine-tuning Settings.} Our fine-tuning configuration follows the one used for linear-probing. In all cases, we use SGD with Nesterov momentum using a momentum parameter of $0.9$. Training uses an initial learning rate of $0.2$ and no weight decay. We use a cosine scheduler with no warmup epochs and a decay factor of $10^{-6}$. Similarly to previous work, for datasets including a validation split, we fine-tune the model on the training and validation splits together, and evaluate on the testing set.

The number of training iterations and data augmentation depend on the target dataset, and are identical to those used for linear probing. Note that we do not run a hyperparameter sweep for selecting either the weight decay or initial learning rate, \ie, these remain fixed for all experiments.

\subsection{Additional MoCo v2 Results}
\label{sec:supp_additional_mocov2_results}

We complement the MoCo v2 results reported in Tables \ref{tab:downstream_performance} and \ref{tab:transfer_learning_full} by computing both their mean and standard deviation over five random trials.

{\noindent \bf Downstream performance on ImageNet.} Table \ref{tab:supp_downstream_performance_mocov2} reports the downstream performance of our SASSL + MoCo v2 representation model, pretrained and linearly probed on ImageNet. Top-1 and top-5 accuracy is computed over five random trials and reported in terms of their mean and standard deviation.

Results show SASSL pretraining and subsequent linear probing on ImageNet yield a notable boost in top-1 classification accuracy, exceeding the default model's performance by over two standard deviations. This statistically significant improvement underscores the efficacy of incorporating SASSL augmentation into the self-supervised learning process.

{\noindent \bf Transfer learning performance.} Tables \ref{tab:supp_transfer_learning01} and \ref{tab:supp_transfer_learning02} show the transfer learning performance of our SASSL + MoCo v2 model. Linear probing and fine-tuning top-1 accuracy is computed over five random trials and reported in terms of its mean and standard deviation.

Both linear probing and fine-tuning benefit from SASSL pretraining. Notably, linear probing achieves significant gains of up to $10\%$, surpassing default performance by over one standard deviation in most cases, although some target datasets show high variations. While fine-tuning exhibits a smaller improvement gap compared to default models, SASSL representation models consistently outperform baselines by up to $6\%$ in average top-1 accuracy, showcasing their robustness across diverse datasets.

Additionally, the performance differences between different choices of style dataset are generally comparable to the measurement uncertainty, which is typically on the order of 0.3 percentage points.  This suggests that the choice of style dataset does not appear to have as significant of an impact as may be expected.  It seems like the main benefit is drawn from using a different style rather than due to anything about the style itself.  We also show in Section~\ref{sec:supp_additional_ablation} that using novel styles drawn from natural images does provide a statistically significant improvement compared to synthetic styles.

It is important to highlight that this trend of shrinking improvement gaps between fine-tuning and linear probing occurs with other SSL methods as well. This is because fine-tuning adjusts the entire model to the target dataset, making pretrained weights act mainly as a refined model initialization.

\begin{table*}
\centering
\small
\caption{\textbf{Additional experiments on transfer learning.} Downstream top-1 classification accuracy of ResNet-50 pretrained via MoCo v2 $+$ SASSL on ImageNet. Accuracy evaluated on six out of twelve target datasets. SASSL generates specialized representations that improve transfer learning performance, both in linear probing and fine-tuning. Mean and standard deviation reported over five random trials.}
\resizebox{\linewidth}{!}{%
\begin{tabular}{cccccccc}
\toprule
& & \multicolumn{5}{c}{\textbf{Target Dataset}}\\
& & CIFAR10 & CIFAR100 & SUN397 & Cars & Caltech-101 & Flowers\\
\midrule
& & \multicolumn{5}{c}{\emph{Linear Probing}}\\
\cmidrule{2-8}
& \makecell{None (Default)}  & $ 89.94\pm 0.24$ & $ 71.93 \pm 0.48$ & $ 69.96\pm 0.33$ & $ 53.15\pm 0.48$ & $ 88.19\pm 0.75$ & $ 93.39\pm 0.58$\\
  \rowcolor{bg_blue}\cellcolor{white}& \textcolor{black}{ImageNet (\textbf{Ours})} & $ 90.93\pm 0.28$ & $73.26 \pm 0.32$ & $69.67 \pm 0.26$ & $\bm{64.87\pm 1.03}$ & $89.3 \pm 0.23$ & $95.27 \pm 0.4$\\
  \rowcolor{bg_blue}\cellcolor{white}& \textcolor{black}{iNat21 (\textbf{Ours})} & $\bm{91.04\pm 0.16}$ & $ 73.29\pm 0.28$ & $\bm{70.07\pm 0.37}$ & $ 63.96\pm 1.19$ & $\bm{89.89 \pm 1.07}$ & $94.7 \pm 0.87$\\
  \rowcolor{bg_blue}\cellcolor{white}& \textcolor{black}{Retinopathy (\textbf{Ours})} & $90.8 \pm 0.22$ & $73.3 \pm 0.38$ & $69.63 \pm 0.55$ & $64.06 \pm 0.79$ & $89.17 \pm 0.28$ & $ 94.94\pm 0.85$\\
  \rowcolor{bg_blue}\cellcolor{white}& \textcolor{black}{DTD (\textbf{Ours})} & $\bf{91.04 \pm 0.2}$ & $\bm{73.41 \pm 0.23}$ & $69.71 \pm 0.44$ & $64.58 \pm 0.71$ & $89.3 \pm 0.26$ & $95.24 \pm 0.22$\\
  \rowcolor{bg_blue}\cellcolor{white}& \textcolor{black}{PBN (\textbf{Ours})} & $90.85 \pm 0.17$ & $73.38 \pm 0.22$ & $69.69 \pm 0.26$ & $64.12 \pm 0.95$ & $89.59 \pm 0.68$ & $\bm{95.45 \pm 0.34}$\\
\cmidrule{2-8}
& & \multicolumn{5}{c}{\emph{Fine-tuning}}\\
\cmidrule{2-8}
& \makecell{None (Default)}  & $96.91 \pm 0.12$ & $83.4 \pm 0.35$ & $74.25 \pm 0.13$ & $83.63 \pm 7.39$ & $89.27 \pm 0.15$ & $95.75 \pm 0.24$\\
  \rowcolor{bg_blue}\cellcolor{white}& \textcolor{black}{ImageNet (\textbf{Ours})} & $ 97\pm 0.09$ & $83.21 \pm 0.18$ & $73.89 \pm 0.39$ & $\bm{90.33 \pm 0.36}$ & $88.26 \pm 0.35$ & $\bm{96.6 \pm 0.14}$\\
  \rowcolor{bg_blue}\cellcolor{white}& \textcolor{black}{iNat21 (\textbf{Ours})} & $\bm{97.05 \pm 0.14}$ & $83.29 \pm 0.27$ & $74.05 \pm 0.3$ & $90.04 \pm 0.39$ & $88.55 \pm 0.48$ & $95.76 \pm 1.75$\\
  \rowcolor{bg_blue}\cellcolor{white}& \textcolor{black}{Retinopathy (\textbf{Ours})} & $96.97 \pm 0.09$ & $\bm{83.68 \pm 0.25}$ & $\bm{74.26 \pm 0.12}$ & $89.96 \pm 0.52$ & $ 88.44\pm 0.44$ & $96.34 \pm 0.28$\\
  \rowcolor{bg_blue}\cellcolor{white}& \textcolor{black}{DTD (\textbf{Ours})} & $96.77 \pm 0.41$ & $83.28 \pm 0.79$ & $74.17 \pm 0.42$ & $89.59 \pm 0.59$ & $\bm{89.54 \pm 1.94}$ & $95.59 \pm 1.61$\\
  \rowcolor{bg_blue}\cellcolor{white} \multirow{-14}{*}{\rotatebox[origin=c]{90}{\textbf{Style Dataset}}} & \textcolor{black}{PBN (\textbf{Ours})} & $96.97 \pm 0.11$ & $83.36 \pm 0.21$ & $74.18 \pm 0.46$ & $89.75 \pm 0.74$ & $88.97 \pm 0.78$ & $95.77 \pm 1.78$\\
\bottomrule
\end{tabular}}
\label{tab:supp_transfer_learning02}
\end{table*}

\subsection{Additional Ablation Studies}
\label{sec:supp_additional_ablation}
SASSL leverages NST to augment pre-training datasets within SSL methods. The proposed technique maintains semantic content by applying transformations solely to the image's texture. Given a pre-training sample, SASSL treats it as the content image and employs established NST techniques to match the distribution of its low-level features to a chosen style reference. This process creates a new image where the scene's objects, represented by high-level features \citep{johnson_2016_perceptual, zhang_2018_unreasonable}, are preserved while the image's texture, represented by the distribution of low-level features \citep{portilla_2000_parametric, zhu_2000_exploring,heeger_1995_pyramid}, aligns with the provided style image.

The following section presents a comprehensive exploration of how the properties of the style reference affect SASSL's generalization to downstream applications.

{\bf \noindent Effect of the Style Representation in Downstream Performance.}
We conduct additional experiments to better understand the effect of the style representation $\hat{\vz}$ in the downstream task performance of models pretrained via SASSL. Specifically, we replace the style latent code, originally taken from a style image, by (i) i.i.d. Gaussian noise $\hat{\vz}\sim \gN(\bm{\mu},\bm{\Sigma})$, and (ii) the style representation of the content image $\hat{\vz}=\vz_{c}$. Note that latter case is equivalent to using the stylization model $\gT$ as an autoencoder, since no external style is imposed over the feature maps of the content image.

In both cases, we pretrain and linearly probe a ResNet-50 backbone on ImageNet using MoCo v2 equipped with SASSL. All training and downstream task settings follow our default configurations, as covered in the Appendix Section \ref{sec:supp_additional_experimental_details} and \ref{sec:supp_downstream_settings}. We also use the recommended blending and interpolation factors $\alpha, \beta \sim \gU(0.1, 0.3)$.

Table \ref{tab:style_ablation} shows the linear probing performance obtained by the two scenarios of interest. As reference, we also include the performance of MoCo v2 with the default data augmentation, as well as MoCo v2 via SASSL using an external style dataset (Painter by Numbers). Results show that using noise as style representation boosts top-1 accuracy by $0.24\%$ with respect to the default data augmentation, while using the content as style reference improves performance by $0.8\%$. This implies that using noise as style representation hinders performance with respect to just encoding and decoding the input image via the stylization network $\mathcal{T}$. On the other hand, using an external style dataset boosts up performance by $2.4\%$, which is a significantly larger improvement over the two scenarios of interest.

Results suggest that the style reference has a strong effect on the downstream performance of the pretrained models. Either by replacing the latent representation of a style image by noise or removing the style alignment process and keeping the compression induced by the Stylization network $\gT$ (by forcing $\hat{\vz}=\vz_{c}$), the improvement provided via Style Transfer data augmentation is significantly smaller than that obtained with our full technique using external style images.

The combined insights from our ablation study and those in \Secref{sec:ablation_study} demonstrate that simply incorporating NST into standard augmentations cannot fully account for the observed accuracy gains. Our findings suggest the existence of additional mechanisms that contribute to the delicate balance between the standard augmentation pipeline and NST. These include feature blending, pixel interpolation, feature maps to align, and style references.

\begin{table*}
\centering
\small
\centering
\caption{\textbf{Effect of the style representation in downstream performance.} Different style representation settings are analyzed when pretraining a ResNet-50 backbone via MoCo v2. Performance reported on a single random trial.}
\begin{tabular}{ccccc}
\toprule
Augmentation & Configuration & Style Dataset & Top-1 Acc. ($\%$) & Top-5 Acc. ($\%$)\\ 
\midrule
  MoCo v2 (Default)  & $-$ & $-$ & $72.97$ & $90.86$\\
\midrule
  \multirow{5}{*}{\makecell{SASSL + MoCo v2\\\textbf{(Ours)}}} & \multirow{5}{*}{\makecell{Probability: $p=0.8,$\\ Blending: $\alpha \in [0.1, 0.3]$\\ Interpolation $\beta \in [0.1, 0.3]$}} & \makecell{Gaussian Noise\\$\hat{\vz}\sim \mathcal{N}(\bm{\mu},\bm{\Sigma})$} & $73.21$ & $91.17$\\
  \cmidrule{3-5}
  &  & $\hat{\vz}=\vz_{c}$ & $73.77$ & $91.64$\\
  \cmidrule{3-5}& & \cellcolor{bg_blue}\Gape[0pt][2pt]{\makecell{PBN\\(\emph{external})}} & \cellcolor{bg_blue}$\bf 75.38$ & \cellcolor{bg_blue}$\bf 92.21$\\
\bottomrule
\end{tabular}
\label{tab:style_ablation}
\end{table*}

\subsection{Computational Requirements}
\label{sec:supp_computational_requirements}

We conducted additional experiments to compare the runtime of our proposed method against the default augmentation pipeline. We measured the \textit{throughput} (augmented images per second) of SASSL relative to MoCo v2's data augmentation. The throughput was calculated by averaging 100 independent runs on $128 \times 128$-pixel images with a batch size of $2,048$. We also report the relative change, which indicates the percentage decrease in throughput compared to the default data augmentation. All experiments were carried out on a single TPU.

Table \ref{tab:supp_computational_requirements} summarizes the throughput comparison. SASSL reduces throughput by approximately $20\%$ due to the computational overhead of stylizing large batches, which involves running a forward pass of the NST model. However, empirical evidence shows that our approach achieves up to a $2\%$ top-1 classification accuracy improvement on multiple SSL techniques. Based on these findings, we consider that SASSL achieves a favorable trade-off between performance and execution time.

\begin{table}[t]
\centering
\rowcolors{2}{bg_blue}{}
\small
\centering
\caption{\textbf{SASSL runtime}. Comparison of the throughput (processed images/second) of SASSL + MoCo v2's data augmentation pipeline vs. the default MoCo v2's pipeline.}
\begin{tabular}{cccccc}
\toprule
Method & Throughput (images/second) & Relative Change $(\%)$\\ 
\midrule
    \makecell{MoCo v2 (Default)} & $37.45$ & $-$\\
    \makecell{SASSL $+$ MoCo v2 \textbf{(Ours)}} & $29.48$ & $21.28$\\
\bottomrule
\end{tabular}
\label{tab:supp_computational_requirements}
\end{table}

\end{document}